\documentclass{article}

\usepackage{arxiv}
\usepackage[utf8]{inputenc} % allow utf-8 input
\usepackage[T1]{fontenc}    % use 8-bit T1 fonts
\usepackage{url}            % simple URL typesetting
\usepackage{booktabs}       % professional-quality tables
\usepackage{amsfonts}       % blackboard math symbols
\usepackage{nicefrac}       % compact symbols for 1/2, etc.
\usepackage{microtype}      % microtypography
\usepackage{lipsum}
\usepackage{graphicx}
\usepackage{stackrel}
\usepackage{xcolor}
\usepackage{makecell}
\usepackage{colortbl}
\usepackage{hyperref}
\usepackage{mathtools}
\usepackage{eucal}
\usepackage{multirow}
\usepackage{array}
\usepackage{lscape}
\usepackage{tabularx}
\usepackage{amsmath}
\usepackage{amssymb}
\usepackage{soul}
\usepackage[ruled,vlined]{algorithm2e}
\usepackage{algpseudocode}
\graphicspath{ {./images/} }
\usepackage{graphicx}
\newcommand{\tuple}[1]{\ensuremath{\langle #1} \rangle}
\algnewcommand{\LeftComment}[1]{\textcolor{gray}{\(\triangleright\) #1}}

\newcommand\emoji[2]{\includegraphics[width = #2pt]{#1}}

\newcommand\exampleend[1]{
\begin{example}
#1
\begin{flushright}
$\square$
\end{flushright}
\end{example}
}

\newtheorem{indicator}{Indicator}
\newtheorem{example}{Example}
\newtheorem{definition}{Definition}
\definecolor{home}{RGB}{49, 121, 118}
\definecolor{work}{RGB}{70, 41, 172}
\definecolor{study}{RGB}{5, 115, 175}
\definecolor{shop}{RGB}{255, 216, 0}
\definecolor{care}{RGB}{220, 52, 57}
\definecolor{leisure}{RGB}{239, 97, 54}
\definecolor{commute}{RGB}{153, 95, 48}
\definecolor{other}{RGB} {167, 64, 154}
\definecolor{smooth}{RGB}{192, 192, 192}
\definecolor{motor}{RGB}{83, 103, 120}
\definecolor{public}{RGB}{121, 154, 184}
\definecolor{other_mode}{RGB}{121, 121, 121}

\begin{document}

\title{Methodology for Mining, Discovering and Analyzing Semantic Human Mobility Behaviors}

\author{
 Clement Moreau, Thomas Devogele, Veronika Peralta, Cyril de Runz \\
  University of Tours, France\\
  \texttt{firstname.lastname@univ-tours.fr} \\
  %% examples of more authors

  \And
 Laurent Etienne \\
  LABISEN, Yncrea Ouest, France\\
  \texttt{laurent.etienne@yncrea.fr} \\
  %% \AND
  %% Coauthor \\
  %% Affiliation \\
  %% Address \\
  %% \texttt{email} \\
  %% \And
  %% Coauthor \\
  %% Affiliation \\
  %% Address \\
  %% \texttt{email} \\
  %% \And
  %% Coauthor \\
  %% Affiliation \\
  %% Address \\
  %% \texttt{email} \\
}

\maketitle
\begin{abstract}
Various institutes produce large semantic datasets containing information regarding daily activities and human mobility. The analysis and understanding of such data are crucial for urban planning, socio-psychology, political sciences, and epidemiology. However, none of the typical data mining processes have been customized for the thorough analysis of semantic mobility sequences to translate data into understandable behaviors. Based on an extended literature review, we propose a novel methodological pipeline called \textsc{simba} (Semantic Indicators for Mobility and Behavior Analysis), for mining and analyzing semantic mobility sequences to identify coherent information and human behaviors. A framework for semantic sequence mobility analysis and clustering explicability based on integrating different complementary statistical indicators and visual tools is implemented. To validate this methodology, we used a large set of real daily mobility sequences obtained from a household travel survey. Complementary knowledge is automatically discovered in the proposed method.
\end{abstract}

\section{Introduction}
\label{intro}

It is becoming increasingly important to have a good understanding of human mobility patterns in many fields, such as transportation \cite{Batty13}, social and political sciences \cite{Lind07,Castellano09}, epidemiology \cite{Pastor15,Chinazzi20}, and smart city design \cite{Pan13}. For the latter, the ability to model urban daily activities correctly for traffic control, energy consumption, and urban planning  \cite{Barthelemy19} has a critical impact on human quality of life and the everyday functioning of cities. To inform policy makers regarding important projects, such as planning new metro lines, managing traffic demand during large events, or constructing shopping malls, we require reliable models of urban travel demand. Such models can be constructed from censuses, household travel surveys, or simulations that attempt to learn about human behaviors in cities using data collected from location-aware technologies \cite{Jiang16,Pappalardo18}. The development of generative algorithms that can reproduce and aid in understanding human mobility behaviors accurately is fundamental for designing smarter and more sustainable infrastructures, economies, services, and cities \cite{Batty12}. 
Typically, mobility analysis focuses on spatiotemporal analysis and the properties of human movement \cite{Barbosa18}. Pioneering works have highlighted the characteristics, regularities, and predictability of human mobility \cite{Barabasi05,Brockmann06,Gonzalez08,Song10a,Song10b,Jiang12}. 

Recently, a major challenge in machine learning and clustering methods has been the ability to explain models both for both practical and ethical purposes \cite{Guidotti19}. Explanation technologies and techniques are immensely helpful for companies that wish to improve the management and understanding of customer needs. They are also important for improving the openness of scientific discovery and progress of research. The need for clear and interpretable results and models is increasing, particularly for black-box algorithms, where data are huge and complex, and for methods with many parameters. Interpretability is crucial for testing, observing, and understanding the differences between models. Therefore, data comprehension also enhances the learning and/or exploration process in terms of validity. 

In this paper, we propose transposing studies on human movement analysis into the semantic domain to learn and understand human activities. We focus on the analysis of semantic sequences of daily activities and attempt to learn and understand the properties of semantic mobility to extract consistent behaviors from a real human mobility dataset. In summary, this paper provides the following main contributions: 

\begin{itemize}
    \item A methodological pipeline called \textit{Semantic Indicators for Mobility and Behavior Analysis} (\textsc{simba}) is proposed to extract, analyze, and summarize coherent behaviors in semantic sequences of mobility data.

    \item We propose a framework for semantic sequence mobility analysis and clustering explicability integrating state-of-the-art indicators.
  
    \item A case study, from a real-world dataset to the extraction of understandable behaviors, illustrating the applicability of our proposal. 
    
\end{itemize}
To the best of our knowledge, such a methodology for mining and interpreting clusters of human semantic mobility behaviors has not been proposed previously. Additionally, our methodology is generic and can be applied to any type of data representing a sequence of semantic symbols (e.g., activities, points of interest (POIs), web pages, and music in playlists). The remainder of this paper is organized as follows. Section \ref{sec:related_work} presents related work on human mobility indicators and methods for behavior extraction.
In Section \ref{sec:methodology}, we introduce some preliminary definitions and present an overview of our approach. We then discuss the design, statistics, and analytical methods used for behavior extraction from semantic sequences. 
Section \ref{sec:case_study} is dedicated to data description and global analysis of the target dataset. Section \ref{sec:sem_clust} discusses the extraction and characterization of behaviors using a clustering method and the explicability of discovered patterns. We also discuss results in this section. Finally, Section \ref{sec:conclusion} concludes this article.

\section{Related work}
\label{sec:related_work}

Questions regarding the extraction of mobility behaviors and comprehension of discovered patterns lie on the intersection of three main subjects, namely the study of human mobility properties, methods for comparing two semantic mobility sequences (i.e., two sequences of daily activities), and the explicability and interpretability of abstruse machine learning models. Therefore, in this section, we summarize major studies on human mobility characteristics as a basis for the requirements of similarity measures between mobility sequences. An extensive review of similarity measures and their properties is presented in the second subsection. The third subsection discusses clustering methods based on arbitrary distance matrices for automatically extracting groups of similar individuals. Finally, we discuss tools for human mobility analysis from the literature and commonly used indicators for describing semantic mobility sequences, as well as the explainability of black-box models, to understand and infer behaviors. 

%%%%%%%%%%%%%%%
\subsection{Human mobility properties}
\label{sec:mob_law}
%%%%%%%%%%%%%%%

Numerous studies on human mobility have shown remarkable heterogeneity in travel patterns that coexist with a high degree of predictability \cite{Alessandretti18}. In other words, individuals exhibit a large spectrum of mobility ranges while repeating daily schedules that are dictated by routine. González et al. analyzed a nation-wide mobile phone dataset and found that human trajectories exhibit a high degree of temporal and spatial regularity. Each individual is characterized by a time-independent characteristic travel distance and has a significant probability of returning to a few frequently visited locations  \cite{Gonzalez08}. In particular, the authors highlighted the following points. (i) According to Brockmann et al. \cite{Brockmann06}, the travel distances of individuals follow a power-law behavior distribution. (ii) The radius of gyration of individuals, which represents their characteristic travel distance, follows a truncated power law. Song et al. \cite{Song10a} observed mobile phone data and determined that the waiting times of individuals (i.e., times between two moves) are characterized by a power-law distribution, confirming the results presented by  Barab\'{a}si \cite{Barabasi05}.
Additionally, in \cite{Song10b}, Song et al. analyzed the movements of individuals based on the Lempel-Ziv algorithm \cite{Kontoyiannis98} nd calculated a value of 93\% potential predictability for user mobility, which demonstrates that a significant portion of predictability is encoded in the temporal orders of visitation patterns. Additionally, despite significant differences in travel patterns, the variability in predictability is weak and largely independent of the distances users cover on a regular basis. This study was continued by Texeira et al. \cite{Teixeira19}, who demonstrated that the entropy of a mobility sequence can be estimated using two simple indicators, namely regularity and stationarity, indicating that trivial indicators can capture the complexity of human mobility.

When considering patterns in human mobility, particularly movements within a single day or week, it is essential to distinguish locations based on their importance. As mentioned previously, people exhibit periods of high-frequency trips followed by periods of lower activity and a tendency to return home on a daily basis. Therefore, most daily and weekly trajectories will start and end at the same location  \cite{Barbosa18}. One method for quantifying the importance of locations is to rank locations. In \cite{Song10a}, location ranking was performed for mobile users based on the numbers of times their positions were recorded in the vicinities of the cell towers covering their locations. It was found that visitation frequency follows a Zipf law. Another method of distinguishing locations is to construct individual mobility patterns in the form of a network. Schneider et al. \cite{Schneider13} used data from both mobile phone users and travel survey respondents to construct weekday mobility networks for individuals. These profiles consisted of nodes for updating visited locations and edges for modeling trips between locations. Daily networks were only constructed for weekdays to identify topological patterns in mobility during a typical day. It was determined that approximately 90\% of the recorded trips made by all users could be described using only 17 daily networks. Another important point is that all of the identified networks contained strictly less than seven nodes and most of the networks exhibited oscillations, which are represented by cyclic links between two or more nodes. This result suggests that these motifs represent the underlying regularities in our daily movements and are useful for the accurate modeling and simulation of human mobility patterns. 

%%%%%%%%%%%%%%%
\subsection{Approaches to semantic mobility sequence mining}
%%%%%%%%%%%%%%%

In mobility mining, two main approaches coexist with distinct goals: sequence pattern mining and clustering methods. The former extracts subsequences of frequent items from trajectories \cite{Giannotti07,Zhang14,Wan18,Ferrero20} to represent an aggregated abstraction of many individual trajectories sharing the property of visiting the same sequence of places with similar travel and visit times. The latter constructs clusters of similar sequences by comparing pairs using a similarity measure. Each cluster represents a coherent behavior and shares mobility features according to similarity measure properties.

Although sequence pattern mining methods are efficient for mining regular fragments and are easy to interpret, they are unsuitable for assessing similarities between individuals, meaning they cannot be used to extract representative groups according to their activities accurately. In general, clustering methods are superior for comparison, classification, and grouping tasks. Therefore, in the remainder of this section, we review related work on clustering processes for mining mobility behavior. Specifically, we focus on difficulties and solutions associated with the choice of a similarity metric between semantic mobility sequences.

%%%%%%%%%%%%%%%
\subsubsection{Similarity measures}
\label{sec:sim_measure}
%%%%%%%%%%%%%%%

Many similarity measures have been proposed or adapted for comparing sequences of symbols, specifically spatial trajectories (e.g., Euclidean, LCSS \cite{Vlachos02}, DTW \cite{Keogt05}, EDR \cite{Chen05} and Fréchet \cite{Alt95}). 
Most of these measures have been adapted for semantic human mobility to compare sequences of routine activities or location histories  \cite{Li08,Jiang12,Lv13}. 
Table \ref{tab:description} summarizes the measures reviewed, which can be classified into two broad categories: measures based on counts of different attributes between sequences (Att) and measures based on edit distances, which measure the cost of the operations required to transform one sequence into the other (Edit). 

Because trajectories are complex objects based on their multidimensional aspects, the construction and analysis of similarity measures remains a challenging task, underlying by \cite{Ferrero16}, and few works have successfully handled multiple dimensions.
In \cite{Furtado16} and \cite{Lehmann19}, two similarity measures for multidimensional sequences called MSM and SMSM, respectively, were defined based on the aggregation of matching functions controlled by weighting distances defined for each dimension of a sequence. These multidimensional similarity measures can embed the richness inherent to mobility data, but require many parameters and thresholds for initialization. This complexity makes it difficult to visualize and interpret the resulting similarity scores. 

Most previous proposals focus on unidimensional semantic sequences. One method for comparing semantic sequences is to represent them as vectors. Such a representation is particularly interesting because it allows the use of a whole family of distance measures that are well-defined metrics, such as the inner product and Euclidean distance.  
In \cite{Elzinga15}, Elzinga and Studer represented sequences as vectors in the inner product space and proposed a context metric called SVRspell that focuses on duration and similarity. However, their representation has an exponential space complexity of  $\mathcal{O}(|\Sigma|^n)$ where $|\Sigma|$ is the size of the alphabet of symbols and $n$ is the size of the sequence. 

Jiang et al. also represented daily activity sequences as vectors. They defined the space of an individual’ s daily activity sequence by dividing the 24 h in a day into five minutes intervals and then used the activity in the first minute of every time interval to represent an individual’s activity during that five minutes timeframe. Principal component analysis was then used to extract appropriate Eigen activities and calculate the Euclidean distances between them \cite{Jiang12}. 
It should noted be that, in this previous work, time slots are the kernel level for comparison in the sense of two individuals with same activities but practised at different times will be evaluated as strongly dissimilar. We call this type of approach a \textit{time-structural approach}. It is effective to group individuals based on they allocate time to different activities, but a major problem with this approach is that two trajectories composed of the same activities practiced at different times will have no similarity, resulting in extreme sensitivity to time. 

To overcome this time issue, other studies have reused measures from optimal matching (OM) methods   \cite{Studer16}such as the edit distance family (e.g., Levenshtein), LCSS, and DTW. These methods measure the dissimilarity between two sequences $S_1$ and $S_2$as the minimum total cost of transforming one sequence (e.g., $S_1$) into the other (e.g., $S_2$) using indels (insertions, deletions) or substitutions of symbols. Each of these operations has a cost that can vary with the states involved. In this manner, depending on the choice of costs applied, groups of individuals can be created differently. 

{}%
\begin{table*}
\caption{Description of main similarity measures for semantic sequences}
\label{tab:description}  
\begin{tabular}{m{3.5cm}ccm{10.05cm}}
\hline 
\multirow{2}{*}{{Measure}} & \multicolumn{2}{c}{{Type}} & \multirow{2}{*}{{Description}}\tabularnewline
\cline{2-3} 
 & {Att.} & {Edit.} & \tabularnewline
\hline 
{MSM \cite{Furtado16}, SMSM \cite{Lehmann19}} & {$\times$} &  & {Agregation of maching functions of each dimension.}\tabularnewline
{SVRspell \cite{Elzinga15}} & {$\times$} &  & {Based on number of matching subsequences and weighted by
the length of subsequences involved.}\tabularnewline
{Jiang et al. \cite{Jiang12}} & {$\times$} &  & {Euclidean distance between appropriate eigen activities.}\tabularnewline
{Hamming} & {$\times$} & {} & {Sum of mismatches with similarity between elements.}\tabularnewline
{DHD \cite{Lesnard10}} &  {$\times$}& {} & {Sum of mismatches with positionwise state-dependent
weights.}\tabularnewline
{Levenshtein distance \cite{Levenshtein66,Wagner74}} & \multirow{1}{*}{} & \multirow{1}{*}{{$\times$}} & \multirow{1}{*}{{Minimum sum of edit costs to turn $S_{1}$ into $S_{2}$.}}\tabularnewline
{CED \cite{Moreau19b}} &  & {$\times$} & {OM with costs weighted by edit position and symbols
nearby.}\tabularnewline
{Trate (TDA) \cite{Rohwer05}} &  & {$\times$} & {OM with costs based on transition rates.}\tabularnewline
{OMSlen \cite{Halpin10}} &  & {$\times$} & {OM with costs weighted by symbol length.}\tabularnewline
\hline 
\end{tabular}{\scriptsize \par}
\end{table*}  

%%%%%%%%%%%%%%%
\subsubsection{OM methods, setting cost and mobility behavior}
%%%%%%%%%%%%%%%
\label{sec:editdist}

\begin{table*}
\caption{Properties of main similarity measures for semantic sequences}
\label{tab:properties} 
{}
\begin{tabular}{m{3.5cm}ccccccc}
\hline 
\multirow{2}{*}{{Measure}} & \multicolumn{7}{c}{{Properties}}\tabularnewline
\cline{2-8} 
 & {Metric} & {T. warp} & {Ctxt} & {Permut.} & {Rep.} & {Sim.} & {Multi. dim}\tabularnewline
\hline 
{MSM \cite{Furtado16}, SMSM \cite{Lehmann19}} &  & {$\times$} &  &  &  &  & {$\times$}\tabularnewline
{SVRspell \cite{Elzinga15}} & {$\times$} & {$\times$} & {$\times$} &  & {$\times$} & {$\times$} & \tabularnewline
{Jiang et al. \cite{Jiang12}} & {$\times$} &  &  &  &  &  & \tabularnewline
{Hamming} & {$\times^{\dagger}$} &  &  &  &  & $\times^\ddagger$ & \tabularnewline
{DHD \cite{Lesnard10}} &  &  & {$\times$} &  &  &  & \tabularnewline
{Levenshtein distance \cite{Levenshtein66,Wagner74}} & \multirow{1}{*}{{$\times^{\dagger}$}} & \multirow{1}{*}{{$\times$}} & \multirow{1}{*}{} & \multirow{1}{*}{} & \multirow{1}{*}{} & $\times^\ddagger$ & \multirow{1}{*}{}\tabularnewline
{CED \cite{Moreau19b}} & {$\times^{\dagger}$} & {$\times$} & {$\times$} & {$\times$} & {$\times$} & {$\times$} & \tabularnewline
{Trate (TDA) \cite{Rohwer05}} &  & {$\times$} & $\times$ &  &  & {$\times$} & \tabularnewline
{OMSlen \cite{Halpin10}} & {$\times$} & {$\times$} & {} &  & {$\times$} &  & \tabularnewline
\hline \tabularnewline
\multicolumn{8}{l}{{$\dagger$ Depends if costs fulfil the triangle inequality and/or parameters.}} \\
\multicolumn{8}{l}{{$\ddagger$ By default discrete metric $\rho(x,y)=\begin{cases}
0 & x=y\\
1 & \text{else}
\end{cases}$}}
\end{tabular}{\scriptsize \par}
\end{table*}

\begin{table*}
\caption{Complexity and parameters of main similarity measures for semantic sequences}
\label{tab:complexity} 
{}%
\begin{tabular}{m{3.5cm}cccm{4.55cm}}
\hline 
\multirow{2}{*}{{Measure}} & \multirow{2}{*}{{Complexity}} & \multicolumn{3}{c}{{Parameters}}\tabularnewline
\cline{3-5} 
 &  & {Subs} & {Indels} & {Other}\tabularnewline
\hline 
{MSM \cite{Furtado16}, SMSM \cite{Lehmann19}} & {$\mathcal{O}(n\times p)$} &  &  & {Set of distances $\mathcal{D}$; weight vector $w$; threshold vector $maxDist$}\tabularnewline
{SVRspell \cite{Elzinga15}} & {$\mathcal{O}\left(|\Sigma|^{\max(n,p)}\right)$} &  &  & {Subsequence length weight $a$; symbol duration weight
$b$}\tabularnewline
{Jiang et al. \cite{Jiang12}} & {$\mathcal{O}(n\times p)$} &  &  & {Number of activities}\tabularnewline
{Hamming} & {$\mathcal{O}(n)$} & {Single, User$^\natural$} &  & \tabularnewline
{DHD \cite{Lesnard10}} & {$\mathcal{O}(n)$} & {Data} &  & \tabularnewline
{Levenshtein distance \cite{Levenshtein66,Wagner74}} & \multirow{1}{*}{{$\mathcal{O}(n\times p)$}} & \multirow{1}{*}{{Single, User$^\natural$}} & \multirow{1}{*}{{Single}} & \multirow{1}{*}{}\tabularnewline
{CED \cite{Moreau19b}} & {$\mathcal{O}(n\times p\times\max(n,p))$} & {Ontology} & {Auto} & {Ontology; Context function $f_{k}$; Context weight
$\alpha$}\tabularnewline
{Trate (TDA) \cite{Rohwer05}} & {$\mathcal{O}(n\times p)$} & {Data} & {Single} & {Transition lag $q$}\tabularnewline
{OMSlen \cite{Halpin10}} & {$\mathcal{O}(n\times p)$} & {User} & {Multiple} & {Symbol length weight $h$}\tabularnewline
\hline 
\multicolumn{5}{l}{{$\natural$ If user specifies a similarity measure.}} \\

\end{tabular}{\scriptsize \par}
\end{table*}

A major challenge in OM-based methods is setting operation costs. This is a particularly difficult problem in social science \cite{Abbott00,Hollister09}. There are essentially three main strategies for choosing operation costs: (i) theory-based cost \cite{Studer16}, which determines costs based on theoretical grounds and a priori knowledge; (ii) feature-based cost, which specifies a list of state attributes on which we wish to evaluate the closeness between states using a similarity 
 measure such as the Gower index \cite{Gower71} or Euclidean distance; and (iii) data-driven cost \cite{Rohwer05}, which assigns a cost that is inversely proportional to the transition rates observed in the dataset. A well-known example of the latter strategy is Dynamic Hamming Distance (DHD) \cite{Lesnard10} where the substitution costs at position $t$ are obtained by the transition rates cross-sectionally observed between $t - 1$ and $t$ and between $t$ and $t + 1$.This method is very effective at generating abnormal sequences and outliers. However, based on its construction, DHD has strong time sensitivity and the number of transition rates that must be estimated is very high, potentially leading to overfitting. Finally, there is an additional type of strategy called (iv) ontology-based cost (utilized in \cite{Moreau19b}that is derived from (i) and (ii). This approach infers costs based on taxonomies (or ontologies) and similarity measures in knowledge graphs \cite{Zhu17}. 

An additional difficulty in setting operation costs lies in the context of sequences or, in other words, considering the symbols in the sequences. As pointed out in \cite{Gonzalez08,Song10b,Alessandretti18}, human mobility has a high degree of regularity. Several approaches have been developed to take advantage of this regularity. In \cite{Halpin10}, the OMSlen method was proposed to reduce the cost of operations for repeating symbols, which is particularly useful for mobility sequence mining. Moreau et al. \cite{Moreau19b} proposed reducing the costs of edit operations for symbols that are similar and/or already present in a sequence and close to the edited position. One consequence of this method is that repetitions of nearby similar symbols and permutations have lower costs, making it a \textit{compositional comparison approach}. This method can bring together sequences with similar contents by allowing for some temporal distortions and repetitions.

Based on the information in \cite{Studer16}, a summary of measure properties is presented in Table \ref{tab:properties} The column ``Metric" denotes measures based on mathematical calculations of distances. ``T. warp" denotes measures allowing time warping when comparing sequences. ``Ctxt" denotes measures that consider the context of a sequence to define cost. ``Permutat." indicates that permutations are allowed with a lower cost and ``Rep." indicates that repetitions are cheaper. Finally, ``Sim" denotes measures that consider a similarity function between symbols and ``Multi. dim" denotes measures that handle multidimensional sequences. 

Table \ref{tab:complexity} presents the computational complexity and some details regarding the parameters of each method. In the ``Complexity" column, $n$ and $p$ denote the lengths of compared sequences. It should be noted that for Hamming-family measures, the sequences must have the same length ($n$). The column ``Parameters" contains the necessary tuning parameters and cost strategies for OM measures. In the ``Subst" columns, an entry of ``User" indicates that the costs are set by the user through a theory- or feature-based strategy. An entry of ``Data" denotes a data-driven method and ``Ontology" refers to am ontology-based strategy. Finally, the ``Indels" column indicates whether there is a single state-independent indel cost, denoted as ``Single" state-dependent user-defined indel costs, denoted as ``Multiple" or indel costs that are automatically set by the measure itself, denoted ``Auto".

\subsubsection{Clustering methods}
\label{sec:clustering}

The extraction of behaviors from a dataset is a process that is typically performed using unsupervised machine learning methods. Clustering methods are based on similarity measures such as those described in the previous subsections and are widely used for the discovery of human behaviors, particularly in sequences of mobility data  \cite{Jiang12,Wesolowski12,Pappalardo18}.
  
However, the topologies created by similarity measures for semantic sequences are difficult to apply. In particular, for OM methods, the axioms of the metric spaces are, usually, not hold.  
A pairwise comparison of semantic sequences results in a distance matrix that is used as an input for a clustering process. To the best of our knowledge, the clustering algorithms that are able to handle arbitrary distances (not necessarily metrics) are PAM (or k-medoid) \cite{Park09}, hierarchical clustering \cite{Kaufman09}, density clustering (DBSCAN \cite{Ester96}, OPTICS \cite{Ankerst99}) and spectral clustering \cite{Ng02},  each of which proposes different hypotheses regarding cluster topology.  

According to the similarity measure and representation of sequences, dimensionality reduction methods can be applied to extract primary dimensions \cite{Jiang12}. 
However, commonly used methods such as PCA can only be used for Euclidean spaces in practice. Alternatively, methods such as UMAP \cite{Mcinnes18}, 
facilitate the reduction of a complex topology defined by an arbitrary metric into a low Euclidean space, which facilitates the visualization of clustering results and the use of other clustering methods, such as those requiring a Euclidean space, including k-means. Additionally, UMAP offers superior preservation of the data global structure, fewer hyperparameters for tuning, and better speeds than previous techniques such as t-SNE \cite{Maaten08}.

Therefore, the advantages of these clustering techniques is that they can be used with arbitrary distances, meaning they can be paired with any of the measures discussed in Section \ref{sec:sim_measure} to implement a clustering module. 

%%%%%%%%%%%%%%%
\subsection{Analysis tools supporting mobility mining}
%%%%%%%%%%%%%%%

\begin{table*}
\caption{Indicators for explainability and analysis of semantic mobility sequences and behaviors in a dataset}
\label{tab:indicator} 

%\begin{tabular}{l m{3cm} m{4.5cm}}
\begin{tabular}{lcm{10.5cm}}
\hline 
{Techniques} & {Refs} & {Description}\tabularnewline
\hline 
\hline 
\multicolumn{3}{c}{\emph{\textbf{Frequency distribution}}}\tabularnewline
\hline 
{Length distribution} &  & {Frequency distribution of sequences length in the
dataset.}\tabularnewline
{State distribution} &  & {Frequency distribution for each symbol $x$ in the
sequences in whole dataset.}\tabularnewline
\hline 
\multicolumn{3}{c}{\emph{\textbf{Transition}}}\tabularnewline
\hline 
{Origin-Destination matrix} &  & {Number of transitions from a state (i.e. symbol) $x_{i}$
to $x_{j}$. }\tabularnewline
{Daily pattern} & \cite{Schneider13}  & {Network representation of sequence. Each edge $(x_{i},x_{j})$
represent a transition from $x_{i}$ to $x_{j}$. }\tabularnewline
\hline 
\multicolumn{3}{c}{\emph{\textbf{Disorder}}}\tabularnewline
\hline 
{Entropy} & \cite{Song10a,Kontoyiannis98} & {Level of ``information",
``surprise", or ``uncertainty"
inherent of a variable's possible outcomes. For sequences, the entropy
also consider temporal patterns.}\tabularnewline
{Predictibility} &  \cite{Song10a} & {Probability that an appropriate predictive algorithm
can predict correctly the user\textquoteright s future whereabouts.}\tabularnewline
{Distinct symbols} & \cite{Teixeira19} & {Number of distinct symbols in the sequence.}\tabularnewline
%{Regularity} & \cite{Teixeira19} & {Ratio between the length and the Distinct symbols
%in the sequence.}\tabularnewline
%{Stationarity} & \cite{Teixeira19} & {Mean duration which user stays continuously in the
%same state.}\tabularnewline
\hline 
\multicolumn{3}{c}{\emph{\textbf{Statiscal dependance measures}}}\tabularnewline
\hline 
{Association rules} & \cite{Agrawal93} & {Relation, based on measures of interestingness, between
two or more variables in a dataset.}\tabularnewline
{Pearson residuals} & \cite{Haberman73} & {Measure of the departure of the independence between
two variables.}\tabularnewline
\hline 
\multicolumn{3}{c}{\emph{\textbf{Centrality}}}\tabularnewline
\hline 
{Mode} &  & {Element with the highest frequency in the dataset.}\tabularnewline
{Medoid} &  & {Element which minimizes the distance to all elements
in the dataset.}\tabularnewline
\hline 
\multicolumn{3}{c}{\emph{\textbf{Scattering and outliers}}}\tabularnewline
\hline 
{Diameter and Radius} &  & {Geometrical interpretation of the distances between elements in the dataset. 
%Max distance between: all elements in dataset for the diameter and the medoid other elements for radius.
}\tabularnewline
{Distance distribution} &  & {Distribution of the distance between in the dataset
or subsetof it (e.g. cluster).}\tabularnewline
{Silhouette} & \cite{Rousseeuw87} & {Quality score of clustering. Measures how similar
an object is to its own cluster (cohesion) compared to other clusters
(separation)}\tabularnewline
{UMAP} & \cite{Mcinnes18} & {Dimensional reduction. Visualization of complex elements in 2D Euclidean spaces with a preservation of local topology.}\tabularnewline
\hline 
\end{tabular}
\end{table*}

Data mining and statistical learning techniques are powerful analysis tools for understanding urban mobility \cite{Pan13}. Several works have proposed frameworks and tools for supporting sequence analysis and mobility mining. The most relevant tools are briefly described in this subsection. 

One of the first and well-known frameworks for mobility knowledge discovery is M-Atlas \cite{Gianotti11}, which provides complete functionalities for mobility querying and data mining centered around the concept of trajectories.

Recently, \cite{Pappalardo19} proposed a statistical python library for mobility analysis called \textit{Scikit-Mobility}. Scikit-Mobility enables to load, clean, process and represent mobility data and analyze these by using the common mobility measures. However, to the best of our knowledge, there is no framework oriented semantic mobility mining. Some toolboxes provide partial statistical support for mobility analysis (mainly oriented to spatial mining). One can see \cite{Pebesma18} for a review of R libraries and \cite{Pappalardo19} for Python. 

Based on TraMineR \cite{Traminer11} functionalities, the geovisualization environment 
eSTIMe \cite{Meunin19} allows the representation of semantic daily mobility information with spatio-temporal content.

Despite the availability of such decision support tools and reporting systems, abstracting and analyzing the main characteristics of a group of semantic sequences and explaining why they are clustered together remains a challenge open problem. In particular, while the interpretability and explainability of mining methods are hot research topics, most methods are limited to a specific problem or domain. To the best of our knowledge, there have only been a few studies on providing a methodology for understand mobility mechanisms in clusters of semantic sequences. The most relevant work is that described in  \cite{Jiang12}, where a K-means clustering method based on a time-structured similarity measure was applied to daily mobility sequences. The authors defined eight clusters corresponding to predefined social demographic variables. Cluster analysis was mainly performed based on sequence index plots, state distributions, and the proportion of social demographic characteristics in each cluster. The Silhouette index \cite{Rousseeuw87} was used to control clustering validity. Although this analysis provides a starting point for understanding the typical behaviors within a cluster, it is incomplete and fails to qualify how consistent the elements in a cluster are with the cluster description (e.g., most extreme elements in a cluster and entropy of sequences in a cluster), as well as the topologies formed by mobility sequences (e.g., daily patterns). These aspects of explainability must be retained and enriched to provide a set of indicators allowing us to understand the globality and diversity of all the elements in a cluster, as well as what makes a cluster coherent. 

Techniques that attempt to explain complex machine learning methods are becoming increasingly popular. For example, the LIME technique \cite{Ribeiro16} attempts to explain the predictions and results of black-box machine learning techniques in an interpretable and faithful manner by training an interpretable model based on local results. Similarly, Guidotti and al. \cite{Guidotti19} proposed some techniques and methods such as association and decision rules, and prototype selection elements (e. g., medoids and diameters) to explain black-box systems to make their results more interpretable. In line with these techniques, we believe that the elaboration of indicators is a crucial point for understanding machine learning models. 

Table \ref{tab:indicator} presents a summary of the different indicators used in state-of-the-art methods that can be used to explain semantic mobility sequences. The indicators are structured into categories corresponding to different perspectives of exploring and explaining data. We let  $X=\tuple{x_1,x_2,...,x_n}$ denote a sequence of symbols constructed from an alphabet $\Sigma$ and let $f$ denote the frequency function.

In this paper, we address the problem of knowledge extraction from human activity sequences to develop models of mobility behaviors. To this end, we reuse many of the techniques introduced in this section and propose several new methods to mine and qualify semantic sequences. 
\section{Methodology}
\label{sec:methodology}

This section details the proposed methodology for the analysis of semantic mobility sequences. First, we summarize the methodological pipeline presented this paper, including the nature of the dataset, selected statistical analysis methods, indicators, and clustering process. The second subsection is dedicated to the enrichment and representation of semantic mobility sequences and the third subsection introduces the indicators used for semantic mobility sequence analysis. The fourth subsection discusses the clustering process and corresponding similarity measure, namely Contextual Edit Distance (CED), as well as a hierarchical clustering process. Finally, the fifth subsection describes the methodology used for cluster analysis and the extraction of semantic mobility behaviors. 

%%%%%%%%%%%%%%%
\subsection{The \textsc{simba} methodological pipeline}
%%%%%%%%%%%%%%%

\begin{figure*}
    \centering{
      \includegraphics[width=0.9\textwidth]{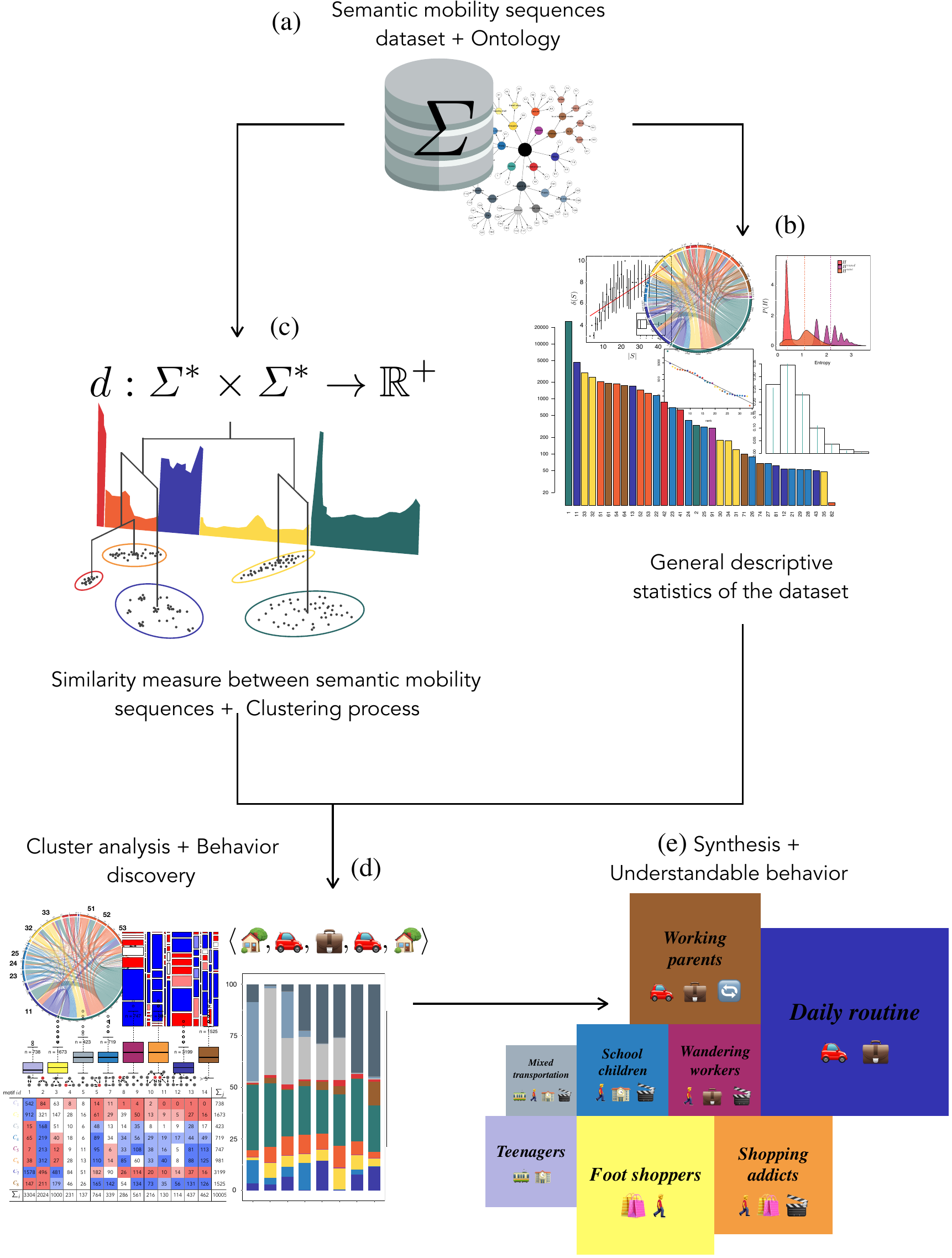}
    }
    \caption{Overview of the simba methodology : (a) Given semantic mobility sequences dataset and an ontology of activities (b) General descriptive statistics and indicators (c) Clustering process (d) Analysis of semantic mobility clusters for behavior discovering (e) Synthesis of each behavior in an understandable way}
    \label{fig:overview}  
\end{figure*}

Semantic mobility sequences are complex data based on their nature and properties, as discussed in Section \ref{sec:related_work}. However, daily mobility has a high degree of regularity with many repetitions of activities. Based on these characteristics, Figure \ref{fig:overview} presents the \textsc{simba} methodology based on the strengths of previous method (also discussed in Section \ref{sec:related_work}). It consists of five steps labeled as (a), (b), (c), (d), and (e). 

In the first step, semantic data are enriched using an ontology to facilitate the comparison of concepts based on any similarity measure that can be adapted to knowledge graphs and data with different levels of granularity.

In the second step, we compute some global statistics to understand and analyze the data. These statistics are selected from the indicators introduced in Table \ref{tab:indicator} and provide a complementary analysis of sequences in terms of their contents (frequency distribution), networks (transition), central behaviors (centrality), and degrees of disorder (entropy). Based on this complementarity, we can explain data from different perspectives. Additionally, the statistics highlight the different patterns that mobility sequences follow (visitation frequency, daily patterns, origin-destination matrices, sequence lengths, predictability) \cite{Song10a,Schneider13,Song10a}), provide a preliminary overview of the data, and facilitate quality control. 

The third step focuses on the clustering of semantic sequences, which groups sequences representing similar moving behaviors. The main challenge in this step is the comparison of semantic sequences, specifically the selection of a similarity measure to support such comparisons and adapt to specific business needs. In this study, we used the CED similarity measure \cite{Moreau19b}. This measure extends edit distance by adapting a cost computation for typical mobility characteristics, such as redundancies, repetitions \cite{Song10b} and cycles \cite{Schneider13}.
A pairwise comparison of semantic sequences yields a distance matrix, which is then used in the clustering process. Section \ref{sec:clustering} summarizes various approaches to sequence clustering. These clustering algorithms are based on different assumptions regarding cluster topology and can all be used in this step. However, because the topology of the semantic sequence space is difficult to comprehend, in this study, we visualized it using a dendrogram generated from a hierarchical clustering process.

The output of this step is a set of clusters of semantic sequences that represent similar behaviors. \textsc{simba} is a modular methodology in which the similarity measure and clustering algorithm proposed in step (c) can be replaced with any of the other techniques discussed in Section \ref{sec:related_work}. 

Step (d) computes additional statistics for each cluster to extract and understand the specific characteristics that constitute mobility behaviors. The statistical and data visualization indicators partially replicate those used in the overall analysis in step (b), but are enhanced with significance tests to determine which are typical characteristics in terms of activities, patterns, and sizes in each cluster. Additionally, these indicators are studied in combination with clustering centrality indicators (centroid, mode, diameter, cluster variance) and quality measures (i.e., Silhouette score \cite{Rousseeuw87}), which measures intra-cluster and inter-cluster distances). This step can also be used to identify outliers. 

Finally, step (e) summarizes the main characteristics of clusters to label them in terms of mobility behaviors. A graphical summary concludes the pipeline and yields an easy and understandable way to discover mobility behaviors. 

The remainder of this section precisely describes each step of the \textsc{simba} methodology.

%%%%%%%%%%%%%%%
\subsection{Enrichment and representation of semantic mobility sequences}
%%%%%%%%%%%%%%%

Let $\Sigma$ be a set of concepts  that represent daily activities (see Section \ref{sec:case_study} and Table \ref{tab:data}). We define semantic mobility sequence as follows.

\begin{definition}[Semantic sequence]

    Given a human $h$, their \textit{semantic sequence}\footnote{Considering our use case, we use indistinctly semantic sequence, semantic mobility sequence or mobility sequence terms.} $S$ is an ordered sequence of activities $\tuple{x_1,x_2,...,x_n}$ such as $\forall k \in [\![1,n]\!], x_k \in \Sigma$ and for $i < j, x_i$ predates $x_j$. 
    Additionally, we consider that symbols are not repeated consecutively (i.e., $\forall k \in [\![1, n-1]\!], x_k \neq x_{k+1}$).
    
    Intuitively, such a sequence indicates that h performed activity $x_1$, then $x_2$, and finally $x_n$. 

\end{definition}

To compare the symbols in $\Sigma$, we must introduce a partial order to the set. For this purpose, we construct a knowledge graph between all concepts in $\Sigma$. 

\begin{definition}[Knowledge graph]
\label{def:ontology}
    Let $\Sigma$ be a set of concepts such that $\exists \mathsf{root} \in \Sigma$. A \textit{knowledge graph} is a connected and directed acyclic graph  $G=(\Sigma,E)$ with $E \subset \Sigma \times \Sigma$ where $(x,y) \in E$ iff the concept $x$ (meronym) \textit{contains} semantically the concept $y$ (holonym) and $\forall (x,y) \in E, y \neq \mathsf{root}$. 
    Such a knowledge graph is called a \textit{meronymy}. 
    
    In the resulting graph of concepts, for any two concepts $x, y\in \Sigma$, we let $LCA(x,y)$ denote the last common ancestor of $x$ and $y$. $d(x)$ denotes for $x$'s depth (i.e., its minimal distance from the \textsf{root} node).
\end{definition}

Additionally, knowledge representations such as an \textit{is-a} taxonomy can induce a partial order on $\Sigma$. This classification of $\Sigma$
allows us to define similarity measures on its elements. Many similarity measures have been proposed for knowledge graphs (see \cite{Zhu17} for a survey).
. In the remainder of this paper, we use the Wu-Palmer similarity measure \cite{Wu94}, which is defined a $sim_{WP}:\Sigma \times \Sigma \rightarrow [0,1]$. This is a well-established state-of-the-art measure that accounts for both the depth of the concepts in an ontology and their closest ancestors, and is normalized:

\begin{equation}
\label{eq:wu-palmer}
    sim_{WP}(x,y) = \frac{2\times d(LCA(x,y))}{d(x)+d(y)}
\end{equation}

Moreover, thanks to the hierarchical representation of activities, data can be analyzed at different aggregation levels, similar to online analytical processing analysis. For example, the activities of shopping in a mall and shopping in a marketplace can be aggregated into a single higher-level shopping activity. Intermediate nodes in a meronymy are useful for such aggregations. 

%%%%%%%%%%%%%%%
\subsection{Semantic mobility sequence dataset analysis}
%%%%%%%%%%%%%%%
\label{sec:indicator}

Semantic sequence data are difficult to analyze based on their combination of temporal dimensions (i.e., temporal order of activities) and semantic dimensions. As discussed in Section \ref{sec:mob_law}, human mobility semantic sequences tend to follow statistical laws. Frequent visitation items induce repetitions of activities that can be modeled using a Zipf law. Sequences are mainly structured by a few networks of daily patterns and are characterized by low entropy and a Poisson distribution for their length. 

\renewcommand{\arraystretch}{1.1}
\begin{table*}
\caption{Retained indicators for semantic mobility sequences analysis}
\label{tab:chosen_indic} 
\begin{tabular}{cllccc}
\hline 
\multirow{2}{*}{{Id}} & \multirow{2}{*}{{Techniques}} & \multirow{2}{*}{{Visualization methods}} & \multicolumn{2}{c}{{Used for}} & \multirow{2}{*}{{Example}}\tabularnewline
\cline{4-5} 
 &  & & {All dataset} & {Clusters} & \tabularnewline
\hline 
\hline 
\multicolumn{6}{c}{\emph{\textbf{Frequency distribution}}}\tabularnewline
\hline 
1 & {Length distribution} &  {Histogram, box plot} & {$\times$} & {$\times$} & Figs. \ref{fig:poisson_length},  \ref{fig:size_clust} \tabularnewline
2 & {State distribution} & {Histogram, stack plot} & {$\times$} & {$\times$} & Figs. \ref{fig:stop_freq}, \ref{fig:stack} \tabularnewline
\hline 
\multicolumn{6}{c}{\emph{\textbf{Transitions}}}\tabularnewline
\hline 
3 & {Origin-Destination matrix} & {Chord diagram} & {$\times$} & {$\times$} & Fig. \ref{fig:flow} \tabularnewline
4 & {Daily patterns} & {Network and histogram} & {$\times$} & {$\times$} & Fig. \ref{fig:daily_patt} \tabularnewline
\hline 
\multicolumn{6}{c}{\emph{\textbf{Disorder}}}\tabularnewline
\hline 
5 & {Entropy} & {Density plot} & {$\times$} & &  Fig. \ref{fig:entropy} \tabularnewline
6 & {Predictability} &  {Density plot} & {$\times$} &  & Fig. \ref{fig:entropy} \tabularnewline
7 & {Distinct symbols} & {Box plot} &  {$\times$} &  &  Fig. \ref{fig:delta} \tabularnewline
\hline 
\multicolumn{6}{c}{\emph{\textbf{Statiscal dependance measures}}}\tabularnewline
\hline 
%8 & {Association rules} & {Table} & $\times$ &  & Tab. ? \tabularnewline
8 & {Pearson residuals} & {Mosaic diagram} &  & {$\times$} & Fig. \ref{fig:mosaic} \tabularnewline
\hline 
\multicolumn{6}{c}{\emph{\textbf{Centrality}}}\tabularnewline
\hline 
9 & {Mode} & {Emojis sequence} &  & {$\times$} & Tab. \ref{tab:centrality} \tabularnewline
10 & {Medoid} & {Emojis sequence} &  & {$\times$} & Tab. \ref{tab:centrality} \tabularnewline
\hline 
\multicolumn{6}{c}{\emph{\textbf{Scattering and outliers}}}\tabularnewline
\hline 
11 & {Diameter and Radius} & {Table} &  & {$\times$} & Tab. \ref{tab:clusters} \tabularnewline
12 & {Silhouette} & {Table} &  & {$\times$} & Tab. \ref{tab:clusters} \tabularnewline
\hline 
\end{tabular}
\end{table*}

Therefore, to ensure the quality of a dataset in terms of the aforementioned properties and obtain a preliminary understanding the data, based on Table \ref{tab:indicator}, we propose complementary statistical indicators that facilitate the global analysis of a set of semantic sequences. The selected indicators are listed in Table \ref{tab:chosen_indic}. Although this study focused on mobility sequences, the proposed methodology is generic and can be used for analyzing any type of semantic sequence dataset. 

\begin{indicator}[Length distribution]
    \label{indic:length}
    Frequency distribution of sequence length combined with a frequency histogram.
\end{indicator}

\begin{indicator}[State distribution]
\label{indic:state}
    Frequency distribution of activities $x \in \Sigma$ inside  the sequences of dataset  combined with a frequency histogram. Using a log scale may be advisable in the field of human mobility. 
\end{indicator}

Together, these two indicators provide a high-level overview of a sequence’s content and length. However, they provide no information regarding the transitions or motifs in sequences. This analysis can be useful for estimating transition probabilities such as those used in DHD measures or for generating probabilistic models of flows. The resulting matrix can be visualized using a chord diagram. To this end, we incorporate the following additional indicators.

\begin{indicator}[Origin-destination Matrix]
\label{indic:od}
    Matrix $T = \{t_{ij}\}$ in which each line/column represents an activity $x\in \Sigma$. 
    The coefficient $t_{ij}$ represents the number of transitions from activity $x_i$ to activity $x_j$. 
    Such a matrix can be visualized using a chord diagram.
\end{indicator}

\begin{indicator}[Daily pattern]
\label{indic:pattern}
    Frequency distribution of non-isomorph daily pattern graphs \cite{Schneider13}. We compute this indicator using Algorithm \ref{alg:daily_patt}:

    \begin{algorithm}[H]
    \SetAlgoLined
    \KwData{Dataset of semantic sequences $\mathcal{D}$}
    \KwResult{Dictionary $\mathcal{G}$ of non-isomorph daily patterns graphs \\ frequencies}
    $\mathcal{G} \gets \emptyset$ \LeftComment{Dictionary $\mathcal{G}$ where keys are graphs and values are integers} \\
     \LeftComment{Construct the daily pattern graph of each sequence $S\in \mathcal{D}$} \\
     \For{$S \in \mathcal{D}$}{ 
     $V_S \gets \{x|x\in S\}$ \LeftComment{Set of vertices}\\
     $E_S \gets \{(x_i, x_{i+1})|i\in [\![1, |S|-1]\!]\}$ \LeftComment{Set of edges}\\
      $G_S \gets (V_S,E_S)$ \\
      \eIf{$\exists G \in \mathcal{G}.keys() | G \simeq G_S$}{ \LeftComment{If there is already exists a graph $G$ isomorph to $G_S$ in $\mathcal{G}$}\\
       $\mathcal{G}[G] \gets \mathcal{G}[G] + 1$  \LeftComment{Increment the frequency of $G$} \\
      } {
       $\mathcal{G}[G_S] \gets 1$ \LeftComment{Create it in $\mathcal{G}$} \\
     }
     }
     \caption{Daily patterns frequency}
     \label{alg:daily_patt}
    \end{algorithm}

   It should be noted that the isomorphism test for the two graphs $G$ and $G'$ can be implemented using the Nauty algorithm \cite{McKay14}. 
\end{indicator}

Finally, to capture the degree of disorder in sequences and understand how studied sequences are both predictable and varied, we use the following three indicators developed in entropy studies.

\begin{indicator}[Entropy of a sequence]
    \label{indic:entropy}
    The entropy of a sequence is defined in \cite{Song10b}, where several types of entropy are given.
    \begin{itemize}
        \item The random entropy $H^{rand} = \log_2\delta(S)$, where $\delta(S)$ is the number of distinct activities in sequence $S$. 
        
        \item The temporal-uncorrelated entropy $H^{unc} = - \sum_{i=1}^{|S|} p(x_i) \log_2p(x_i)$, where $p(x_i)$ is the historical probability that activity $x_i$ was performed. This characterizes the heterogeneity of activities.
        
        \item The real sequence entropy $H$ which depends on both, the frequency and the order of an activity in the sequence. 
        
    \begin{equation} 
    \label{eq:entropy}
    H(S) = - \sum_{S'\subset S} p(S') \log_2p(S')
    \end{equation} 
    
    where  $p(S')$ is the probability of finding a particular ordered subsequence $S'$ in the sequence $S$.
    \end{itemize}
    
    In practice, $H$ is uncomputable for long sequences. Therefore, we use the following estimator $H^{est}$ of $H$ proposed in \cite{Kontoyiannis98}:
    
    \begin{equation} 
    \label{eq:entropy_est}
    H^{est}(S) = \left(\frac{1}{|S|}\underset{i}{\sum}\lambda_{i}\right)^{-1}\log_{2}|S|
    \end{equation} 
    
    where $\lambda_i = \text{argmin}_{k\geq 1}\{x_i...x_k \notin x_1...x_{i-1}\}$ is the size of the smallest subsequence beginning at $i$ and not contained in the and not contained in the range of 1 to $i-1$.
    
    Kontoyiannis et al. demonstrated that $\lim_{|S| \rightarrow \infty} H^{est}(S) = H(S)$,  supplements can also be derived \cite{Song10b}.

\end{indicator}

\begin{indicator}[Predictability]
    \label{indic:predict}
    The predictability $\Pi$ that an appropriate algorithm can predict correctly the user’s future whereabouts. Thanks to Fano's inequality, we can obtain an upper bound $\Pi^{max}$ for $\Pi$ \cite{Song10b}. $\Pi^{max}$ is obtained via the approximate resolution of the following equation:
    
    \begin{equation}
        H(S) = \mathcal{H}(\Pi^{max}) + \left( 1-\Pi^{max} \right) \log_2(|S|-1) 
        \label{eq:predict}
    \end{equation}
    
    where $\mathcal{H}(x) = -x \log_2 x - (1-x) \log_2(1-x)$ is the binary entropy function.
\end{indicator}

\begin{indicator}[Distinct symbols]
\label{indic:unique}
    The frequency distribution of the number of distinct activities $\delta$ in each sequence $S$ in the dataset combined with a frequency histogram.
    $\delta$ can also be studied in combination with the length $|S|$ to uncover hidden regularities in a sequence. 
\end{indicator}

%%%%%%%%%%%%%%%
\subsection{Clustering design for semantic sequences}
%%%%%%%%%%%%%%%
\label{sec:clustering_process_metho}

To address the problem of clustering semantic mobility behaviors in a metropolitan area utilizing a compositional approach (i.e., ``What does an individual do during a day?"), we use a combination of the CED measure and hierarchical clustering based on Ward's criterion. 

\subsubsection{CED}
\label{sec:ced}

As discussed in Section \ref{sec:editdist}, the distances in the edit distance family count the minimum costs of operations (e.g., modification, addition, deletion) required to transform one sequence into another. Such measures can be used to quantify the similarity between two semantic mobility sequences. However, as indicated in Section \ref{sec:mob_law}, human mobility sequences are characterized by redundancy of certain symbols, repetition \cite{Song10b}, and cycles \cite{Schneider13}. These features should be considered by adopting specialized distances. 

Based on these observations, we proposed the use of the CED measure \cite{Moreau19a,Moreau19b}, which is a generalization of edit distance for handling semantic mobility sequences. This measure incorporates the following factors: 

\begin{enumerate}
    \item \textit{Context-dependent cost}: Edit cost depends on the similarity of nearby activities. The more similar and closer two activities are, the lower the cost of operations..
    \item \textit{Repetition}: Editing repeated nearby activities has a low cost.
    \item \textit{Permutation}: Similar and nearby activities can be exchanged with a low cost.
\end{enumerate}

These three factors of CED are particularly suitable for mobility analysis. The fact that repetition and the editing of similar elements in a sequence has a low cost, similar to permutation, tends to group elements with activities with the same semantics while accounting for a flexible timeframe. 

To achieve these advantages, the CED includes a modification of the cost operation function $\gamma$ that generalizes the classical definition of edit distance and accounts for the local context of each activity in a mobility sequence. 

Let a contextual edit operation  be a quad tuple such that: \[e=(o,S,x,k) \in \{\mathtt{add}, \mathtt{mod}, \mathtt{del}\}\times \Sigma^n \times \Sigma\cup \{\varepsilon\} \times \mathbb{N}^*\]
where $e$ is a transformation $o$ of sequence $S$ at index $k$ using symbol $x$. Let $E$ be the set of all possible contextual edit operations, the cost function $\gamma : E \rightarrow [0,1]$ for a contextual edit operations is defined as: 

\begin{equation}
\label{eq:costFunction}
 \gamma(e)= 
 \alpha \times \ell(e) + \\
 (1- \alpha)\left(1- \underset{i\in [\![1, n]\!]  }{\max}\left\{ sim(x ,s_i)\times v_{i}(e)\right\} \right)
\end{equation}

where: 
\begin{itemize}
    \item $\alpha \in [0,1]$ is a contextual coefficient. \\ If $\alpha \rightarrow 0$, then the cost will be strongly evaluated according to the near content at index $k$ in the sequence being edited. If $\alpha \rightarrow 1$, then CED tends toward the Levenshtein Distance with substitution cost.
    \item $\ell(e)=\begin{cases}
            1-sim(s_k, x) & if\ o = \text{\texttt{mod}}\\
            1 & else
            \end{cases}$ 
            is the cost function of Levenshtein Distance with substitution cost. 
    \item $sim:\Sigma\times \Sigma \rightarrow [0,1]$ 
    is a similarity measure between two activities. 
    \item $v(e)\in [0,1]^n$ is a contextual vector that quantifies the notion of proximity between activities. Typically, the larger $|i-k|$ is, the smaller $v_i(e)$ is. 
\end{itemize}

Let $\mathcal{P}(S_1,S_2)$, all the edit paths to transform a sequence $S_1$ into $S_2$, the one-sided contextual edit distance from $S_1$ to $S_2$ noted $\tilde{d}_{CED}:\Sigma^n \times \Sigma^p \rightarrow \mathbb{R}^+$ is defined such that:

\begin{equation}
\label{eq:one_ced}
    \tilde{d}_{CED}(S_1,S_2) = \underset{P\in\mathcal{P}(S_{1},S_{2})}{\min}\left\{ \stackrel[i=1]{|P|}{\sum}\gamma(e_{i})\right\} 
\end{equation}
where $P=(e_1,...,e_q)\in E^q$ is a vector of contextual edit operations. 

The computation of Equation \ref{eq:one_ced} is performed using dynamic programming and the Wagner-Fisher algorithm \cite{Wagner74}. Finally, $d_{CED}:\Sigma^n \times \Sigma^p \rightarrow \mathbb{R}^+$ is computed using the following equation:

\begin{equation}
\label{eq:ced}
    d_{CED}(S_1,S_2) = \max\left\{\tilde{d}_{CED}(S_1,S_2), \tilde{d}_{CED}(S_2,S_1)\right\} 
\end{equation}

%%%%%%%%%%%%%%%
\subsubsection{Hierarchical clustering settings and validity}
%%%%%%%%%%%%%%%
\label{sec:hierarchical_clust}

Hierarchical clustering algorithms have been widely applied to partition datasets into different clusters \cite{Rokach05}. In the case of an abstract topological space, similar to the space constructed using the CED for semantic mobility sequences, the dendrogram used to visualize the results of hierarchical clustering provides support for understanding the studied space. However, in addition to defining a similarity measure, hierarchical clustering requires three other parameters to be defined: the strategy (top-down or bottom-up), linkage criterion, and dendrogram cutoff method.

Regarding the choice of strategy, the bottom-up approach has a polynomial time complexity of  $\mathcal{O}(n^2\log n)$ versus an exponential complexity of $\mathcal{O}(2^n)$ for the top-down approach \cite{Kaufman09}. Therefore, to handle a large dataset (in our case, 10\! 005 sequences), we used a hierarchical agglomerative clustering (HAC) algorithm based on a bottom-up strategy. A summary of hierarchical clustering algorithms in statistical software is presented in \cite{Struyf97}. 

Regarding linkage criteria, the authors of \cite{Kaufman09} summarized the common options used in the literature. This choice depends on cluster shapes. The simplest approach, namely the single linkage criterion, is based on the minimum distance between a pair of elements from two clusters and can handle any cluster shape. However, repeated merges can lead to a chaining effect. In contrast, complete linkage, which is based on maximum distances, produces more compact clusters, but is sensitive to noise and outliers. Average linkage is particularly useful for convex clusters \cite{Kaufman09}.
Because we do not know the shapes of clusters beforehand and we want robustness to outliers and immunity to chaining effects, we adopted the Ward criteria that minimize the total within-cluster variance. This is similar to the K-means algorithm, which is less affected by noise and tends to create convex compact clusters. 

Finally, the determination of the optimal number of clusters can be considered from different perspectives and is a relatively difficult problem \cite{Halkidi01}. The simplest method for hierarchical clustering is based on higher relative loss of inertia criteria \cite{Krzanowski88} 
This method is associated with the largest gap between two successive agglomerations in a dendrogram. A summary of the different techniques can be found in \cite{Halkidi01}. 

Finally, quality clustering indicators such as the Silhouette score \cite{Rousseeuw87} are useful criteria for assessing the natural number of clusters and ensuring the validity of clustering. In particular, the Silhouette score is based on the same objective function as the Ward criterion and can be maximized to determine the optimal number of clusters.

%%%%%%%%%%%%%%%
\subsection{Analysis of semantic sequence clusters}
%%%%%%%%%%%%%%%
\label{sec:clust_anal_method}

Let $\{C_1, ..., C_m\}$ be a partition of the dataset $\mathcal{D}$ of semantic mobility sequences where $C_{k\in [\![1,m]\!]}$ represents a cluster. In this section of the pipeline,  we wish to extract the meaningful characteristics of each cluster $C_k$ in order to understand and explain the mobility behavior of each cluster. 
To this end, we calculate most of the indicators defined in Section \ref{sec:indicator} for each cluster $C_k$. 

For numerical frequency distribution indicators,  such as Indicator \ref{indic:length} or Indicator \ref{indic:unique}, we use boxplots to summarize and compare the distributions of each cluster. In contrast, for categorical frequency distribution indicators such as Indicator \ref{indic:state} and Indicator \ref{indic:pattern}, according to the process described in \cite{Oliveira03}, we use contingency tables, mosaic plots \cite{Friendly94} and stack plots to visualize information. In this phase, the indicators are enriched with significance tests such as chi-squared test and Pearson residuals \cite{Haberman73} in order to identify under- or over-representation of some variables, patterns or activities in the clusters. Cramér's $V$ score is used to evaluate the strength of relationships between these variables and clusters. 

\begin{indicator}[Pearson residuals]
\label{ind:pearson}
    Consider a sample of size $N$ of the simultaneously distributed variables $A$ and $B$ with $a_1, ..., a_p$ and $b_1, ..., b_q$, let $(n_{ij}),1 \leq i \leq p, 1\leq j\leq q$ be the number of times the values $a_{i}$ and $b_{j}$ are observed, and let $(n^*_{ij}) = \frac{n_{+j} \times n_{i+}}{N}$ be the theoretical values where:
    \begin{itemize}
        \item $n_{+j}=\sum_{i=1}^p n_{ij}$, represents the column marginal for that cell, 
        \item $n_{i+}=\sum_{j=1}^q n_{ij}$ represents the row marginal for that cell.
    \end{itemize}  
    Then, the Pearson residuals $r_{ij}$ \cite{Haberman73} are defined as:
    \begin{equation}
    r_{ij} = \frac{n_{ij} - n^*_{ij}}{\sqrt{n^*_{ij}}}
\end{equation}
  
Pearson residuals represent the strength and direction of the association between $a_i$ and $b_j$. The strength is defined by the absolute value of the residual and the direction by its sign. Units are in standard deviations, meaning a residual greater than 2 or less than -2 represents a significant departure from the independence at the 95\% confidence level. 
\end{indicator}

By calculating Pearson residuals, we can determine how much the observed values deviate from the values in the case of complete independence. For example, an interesting subject is the departure of the frequency values of an activity $x_i\in \Sigma$ in a given cluster $C_j$. If $|r_{ij}|\geq 2$, we can conclude that $x_i$ has a statistically significant association with cluster $C_j$, where the sign indicates if $x_i$ is under- (negative sign) or over- (positive sign) represented in $C_j$. However, statistical significance does not necessarily imply a strong association. 
There is a more standardized strength test called the chi-squared test. Statistical strength tests represent correlation measures. For the chi-squared test, the most commonly used measure is Cramér's $V$ score \cite{Cramer99}. $V$ varies from zero (corresponding to no association between variables) to one (complete association) and can reach one only when each variable is completely determined by the other.

Therefore, these measures can be used to characterize the activities or daily patterns in a cluster and provide partial information regarding the behaviors represented by patterns. However, these significance tests do not provide meaningful information regarding the order in which activities are conducted. The origin-destination matrix provides some additional information regarding the other of activities, but it cannot represent complete coherent behaviors in a cluster. However, indicators of centrality such as the medoid and the mode of a cluster can be used to extract an archetypal mobility sequence from the cluster. 

\begin{indicator}[Mode]
   Given a set of elements $C$ and a similarity measure $d:C\times C \rightarrow \mathbb{R}^+$, the mode $M$ of $C$ is defined such that:
   \begin{equation}
   \label{eq:mode}
       M=\underset{X\in C}{\text{argmax}}\left\{f(X)\right\} 
   \end{equation}
   where $f$ denotes the frequency function. Intuitively, $M$ is the element which is the most-frequent in $C$. 
\end{indicator}

\begin{indicator}[Medoid]
\label{ind:medoid}
   Given a set of elements $C$ and a similarity measure $d:C\times C \rightarrow \mathbb{R}^+$, the medoid $m$ of $C$ is defined such that:
   \begin{equation}
   \label{eq:medoid}
       m=\underset{X\in C}{\text{argmin}}\left\{ \underset{Y\in C}{\sum}d(X,Y)\right\} 
   \end{equation}
   Intuitively, $m$ is the element that minimizes the distance to all other elements in $C$. 
\end{indicator}

To validate whether the medoid $m$ actually represents the elements of a cluster, it is essential to study the topology of the cluster $C$. Here, $m$ is a good representative of $C$ if the formed cluster is hyperspherical (i.e., the distribution of distances $d(x,m)$ follows a power law which indicating that most of elements are near $m$).
Furthermore, hierarchical clustering achieves a complete partitioning of a dataset. Therefore, this analysis can identify outlier elements in clusters which can be considerate as the 5\% of elements far away from the medoid\footnote{Under the hypothesis of hyperspherical clusters}. Another measure for studying scattering and outliers that ignores the topology of clusters is cluster diameter.

\begin{indicator}[Diameter and Radius]
   Given a set of elements $C$ and a similarity measure $d:C\times C \rightarrow \mathbb{R}^+$, the diameter $diam$ of $C$ is defined as:
   \begin{equation}
   \label{eq:diameter}
       diam(C)=\underset{X,Y\in C}{\max}\left\{d(X,Y)\right\}
   \end{equation}
   where $diam$ represents the greatest distance between any pair of elements in the cluster. It should be noted that $diam$ can also represent the most-distant pair of elements if $\max$ is replaced with $\text{argmax}$ is Equation \ref{eq:diameter}. 
   Similarly, the radius $rad$ of $C$ is defined as:
   \begin{equation}
   \label{eq:radius}
       rad(C)=\underset{X\in C}{\max}\left\{d(m,X)\right\}
   \end{equation} 
  where $m$ is the medoid of $C$ such as defined by Indicator \ref{ind:medoid}.
\end{indicator}

Finally, analysis can be completed by calculating the Silhouette score of a cluster. 

\begin{indicator}[Silhouette]
\label{ind:Silhouette}
       Let $\{C_1, ..., C_m\} $ a partition of the dataset $\mathcal{D}$. The Silhouette score \cite{Rousseeuw87} is a value which quantifies how is appropriately $X\in C_k$ is clustered. It is defined as: 
       \begin{equation}
       sil(X) = \frac{b(X)-a(X)}{\max\{a(X), b(X)\}}
       \end{equation}
       where: 
    \begin{itemize}
        \item $a(X) = \frac{1}{|C_k|-1}\underset{Y\in C_k,Y\neq X}{\sum}d(X,Y)$
        \item $b(X)=\underset{C_i\neq C_k}{\min}\frac{1}{|C_i|}\underset{Y\in C_i}{\sum}d(X,Y)$
    \end{itemize}
    
    Here, $a(X)$ is the mean of the distance between $X$ and all other elements in $C_k$. Thefore, it can be interpreted as a measure of how well $X$ is assigned to its $C_k$. On the other hand, $b(X)$ is the smallest mean distance from $X$ to every points in the other clusters. The cluster with the smallest mean dissimilarity is said to be the ``neighboring cluster" of $X$ because it is the next-best fit cluster for point $X$. 
    
    Thus, the Silhouette score of a cluster $C_i$ is defined by the arithmetical mean of $sil(X)$ for each $X\in C_k$:
    
    \begin{equation}
        Sil(C_k) = \frac{1}{|C_k|}\sum_{X\in C_k}sil(X)
    \end{equation}
\end{indicator}

The next section illustrates the application of the proposed methodology to a real-world dataset and describes our findings in terms of mobility behaviors.  

\section{Case study}
\label{sec:case_study}

To test the proposed methodology, we used a set of real mobility sequences obtained from a French household travel survey called EMD.\footnote{from French ``Enquête Ménages-Déplacements"}. 
The goal of the EMD survey is to provide a snapshot of the trips undertaken by residents of a given metropolitan area, which can aid in understanding mobility behaviors and measure changes over time.

In this section, we describe the EMD data in terms of quality, semantics, and size. The dataset is complemented by a domain ontology describing activity semantics (step (a) in the methodology in Fig. \ref{fig:overview}). A statistical study and overview analysis of the dataset conclude this section (step (b) in Fig. \ref{fig:overview}). 

%%%%%%%%%%%%%%%
\subsection{EMD Rennes 2018 dataset}
%%%%%%%%%%%%%%%

\begin{table*}
\caption{Description of activities in the EMD data}
\label{tab:data} 

\begin{tabular}{m{.5cm}>{\centering}m{3.45cm}cm{10cm}}

\hline 
{Color} & {Aggregated activity} & {Emoji} & {Activity label and description}\tabularnewline
\hline 
\hline 
\multicolumn{4}{c}{\emph{Stop activities}}\tabularnewline
\hline 
\hline 
{\cellcolor{home}} & {Home} & \emoji{Figures/Emoji/home}{10} & {\textbf{1}: main home; \textbf{2}: second home, hotel}\tabularnewline
\hline 
{\cellcolor{work}} & {Work} & \emoji{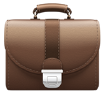}{10} & {\textbf{11}: work in official work place; \textbf{12}: work at home; \textbf{13}:
work in another place; \textbf{43}: look for a job; \textbf{81}: do a work round}\tabularnewline
\hline 
{\cellcolor{study}} & {Study} & \emoji{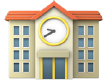}{12} & {\textbf{21}: day nursery; \textbf{22}: study at school (primary); \textbf{23}: study at school (college); \textbf{24}: study at school (high school); \textbf{25}: study at school (university); \textbf{26}: study in another place (primary); \textbf{27}: study in another place (college);
\textbf{28}: study in another place (high school); \textbf{29}: study in another place
(university)}\tabularnewline
\hline 
{\cellcolor{shop}} & {Shopping} & \emoji{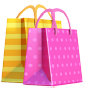}{10} & {\textbf{30}: visit a shop; \textbf{31}: visit a shopping center; \textbf{32}:
shopping in mall; \textbf{33}: shopping in medium or little shops; \textbf{34}:
do shopping in market place; \textbf{35}: drive-through shopping}\tabularnewline
\hline 
{\cellcolor{care}} & {Personal Care} & \emoji{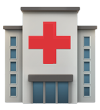}{10} & {\textbf{41}: health care; \textbf{42}: administration step}\tabularnewline
\hline 
{\cellcolor{leisure}} & {Leisure} & \emoji{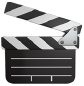}{10} & {\textbf{51}: sport, cultural or voluntary activity; \textbf{52}: go for
a walk or window-shopping; \textbf{53}: go in restaurant; \textbf{54}: visit family
or friend; \textbf{82}: do a shopping tour (more than 4 consecutive activity
30)}\tabularnewline
\hline 
{\cellcolor{commute}} & {Accompany} & \emoji{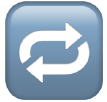}{10} & {\textbf{61}, \textbf{63}: go with someone; \textbf{62}, \textbf{64}: pick-up someone; \textbf{71},
\textbf{73}: drop-off someone to a transport mode; \textbf{72}, \textbf{74}: collect someone
to a transport mode}\tabularnewline
\hline 
{\cellcolor{other}} & {Other} & \emoji{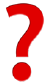}{7} & {\textbf{91}: other (detail in notes)}\tabularnewline
\hline 
\hline 
\multicolumn{4}{c}{\emph{Move activities}}\tabularnewline
\hline 
\hline 
{\cellcolor{smooth}} & {Smooth} & \emoji{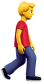}{7} & \textbf{100}{: walk; }\textbf{110}{:
ride location bike; }\textbf{111}{: ride
bike; }\textbf{112}{: bike passenger; }\textbf{193}{:
roller, skateboard, scooter; }\textbf{194}{:
wheelchair; }\textbf{195}{: small electric
machines (electric scooter, segway, etc)}\tabularnewline
\hline 
{\cellcolor{motor}} & {Motorized} & \emoji{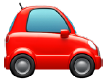}{10} & \textbf{113}{: motor bike driver ($< 50cm^{3}$);
}\textbf{114}{: motor bike passenger ($<
50cm^{3}$); }\textbf{115}{: motor bike
driver ($\geq50cm^{3}$); }\textbf{114}{:
motor bike passenger ($\geq50cm^{3}$); }\textbf{121}{:
car driver; }\textbf{122}{: car passenger;
}\textbf{161}{: taxi passenger; }\textbf{171}{:
car transport (work); }\textbf{181}{: van
or truck driver (for activity 81); }\textbf{182}{:
van or truck driver (for activity 81); }\tabularnewline
\hline 
{\cellcolor{public}} & {Public transportation} & \emoji{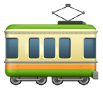}{10} & \textbf{131}{: urban bus passenger; }\textbf{133}{:
subway passenger; }\textbf{13}{8, }\textbf{139}{:
other public transportation passenger; }\textbf{141}{,
}\textbf{142}{: local public transportation;
}\textbf{151}{: train passenger}\tabularnewline
\hline 
{\cellcolor{other_mode}} & {Other mode} & \emoji{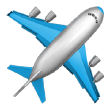}{10} & \textbf{191}{: sea transport; }\textbf{192}{:
airplane; }\textbf{193}{: other modes (agricultural
equipment, quad bike, ect);}\tabularnewline
\hline
\end{tabular}
\end{table*}

The studied dataset is called ``EMD Rennes 2018" and it represents a household travel survey conducted in Rennes city and the surrounding area (Britany region of France). The survey was conducted from January to April of 2018 during weekdays. The data represent 11\! 000 people (at least five years old) from 8\! 000 households. This sample is considered to be statistically representative of 500\! 000 households and one million residents. The details of the data collection methodology and its quality are discussed in \cite{certu08}, and a summary of the results of the EMD Rennes 2018 survey is presented in \cite{audiar19}\footnote{References in french}. 

The dataset consists of a set of mobility sequences, each of which represents the activities performed by one person over 24h. Table \ref{tab:data} lists the different activity labels used in the EMD mobility sequences. Two main classes are represented: stop activities and move activities. The former corresponds to daily static activities such as ``staying at home", ``working" and ``shopping". The latter represents transportation activities such as ``walking" or ``driving a car." 

Mobility sequences are defined based on the Stop-Move paradigm \cite{Parent13}. Each stop activity is followed by one (or more) move activities. Therefore, the time dimension is only considered in terms of the order of the activities, resulting in a \textit{compositional approach} to mobility analysis. 

\exampleend{
\label{ex:seq}

Consider the following activities performed by Sam during a day:
\vspace{.25cm}
\textit{``Sam starts her day at home. Then, she walks to the bus station and takes the bus to work. She spends her work time at her office and then walks home."}
\vspace{.25cm}

The mobility sequence $S$, which is represented below, corresponds to Sam's activities. By using activity codes in Table \ref{tab:data}, we have  $S=\tuple{1, 100, 131, 11, 100, 1}$. 
Alternatively, by considering aggregated activities, represented by emojis, we obtain the following representation: 
$S_{agg}=\langle$\emoji{Figures/Emoji/home}{10}, \emoji{Figures/Emoji/smooth}{6}, \emoji{Figures/Emoji/public}{11},
\emoji{Figures/Emoji/work}{10},
\emoji{Figures/Emoji/smooth}{6}, \emoji{Figures/Emoji/home}{10}$\rangle$. 
}

Throughout this paper, we will use activity codes and emojis to represent sequences both in examples and when analyzing real sequences. Among the 11\! 000 sequences in the dataset (corresponding to 11\! 000 surveyed individuals), we filtered those containing no moves (corresponding to people that stayed at home the entire). This resulted in a final dataset of 10\! 005 mobility sequences. 

\begin{figure*}[t!]
    \includegraphics[width=\textwidth]{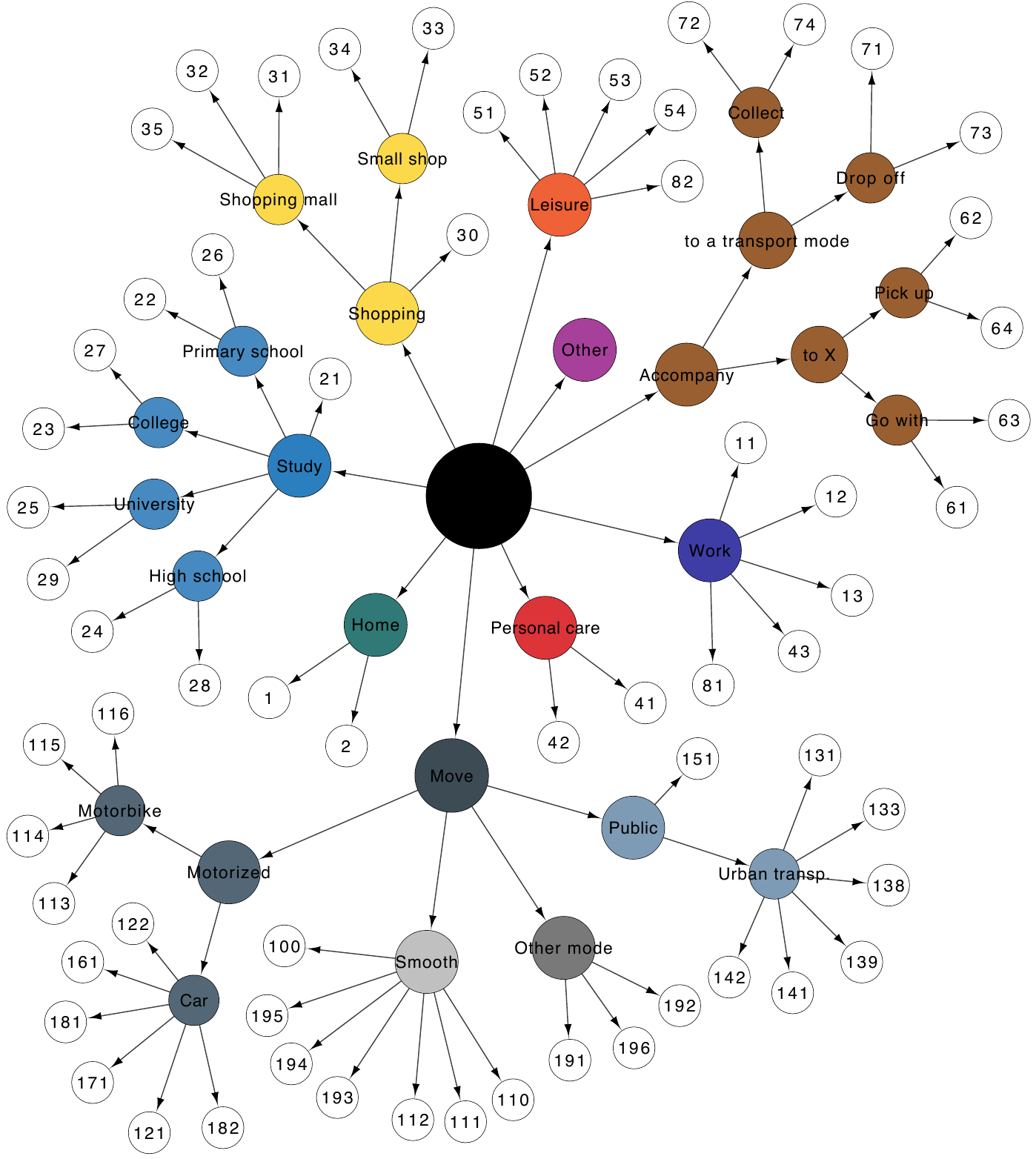}
    \caption{EMD graph ontology}
    \label{fig:ontology}       
\end{figure*}

%%%%%%%%%%%%%%%
\subsection{Ontology} 
%%%%%%%%%%%%%%%

The activity concepts detailed in Table \ref{tab:data} are also structured in a knowledge graph (or ontology), as shown in Fig. \ref{fig:ontology}. This knowledge graph refers to Definition \ref{def:ontology} and is a hybrid of the EMD meronomy and the Harmonised Time Use Surveys (HETUS) \cite{Eurostat19}.
Each color corresponds to a meta-category representing \textit{aggregated activities}. irst-level nodes for stop activities and second-level nodes for move activities (i.e., transport modes). Inter-level nodes come from the HETUS classification and first-level nodes and leaves come from the EMD survey. 

Other possibilities for arranging concepts can be considered, each of which refers to a particular study context or specific business need. The structure of a graph influences the similarity measures between concepts. 

\exampleend{
\label{ex:onto}

    Suppose we wish to compute the similarity between activities \textbf{100} (walking) and \textbf{121} (car driving). Using the ontology in Fig. \ref{fig:ontology}, we can compute the Wu-Palmer similarity defined in Equation \ref{eq:wu-palmer} as:
    
    $\begin{aligned} 
    sim_{WP}(100,121) &= \frac{2\times d(LCA(100, 121))}{d(100)+d(121)} \\ 
            &= \frac{2\times d(\text{Transport mode})}{d(100)+d(121)} \\ 
            &=  \frac{2}{7} \\ 
    \end{aligned}$ 

    where $LCA(x,y)$ is the Last Common Ancestor of concepts $x$ and $y$ and $d(x)$ is the shortest path between node $x$ and the root node (depicted in black in Fig. \ref{fig:ontology}). 
}

%%%%%%%%%%%%%%%
\subsection{Comprehensive statistical analysis of the dataset}
%%%%%%%%%%%%%%%
\label{sec:glob_analysis}

\begin{figure}[p]
    \includegraphics[width=0.85\textwidth]{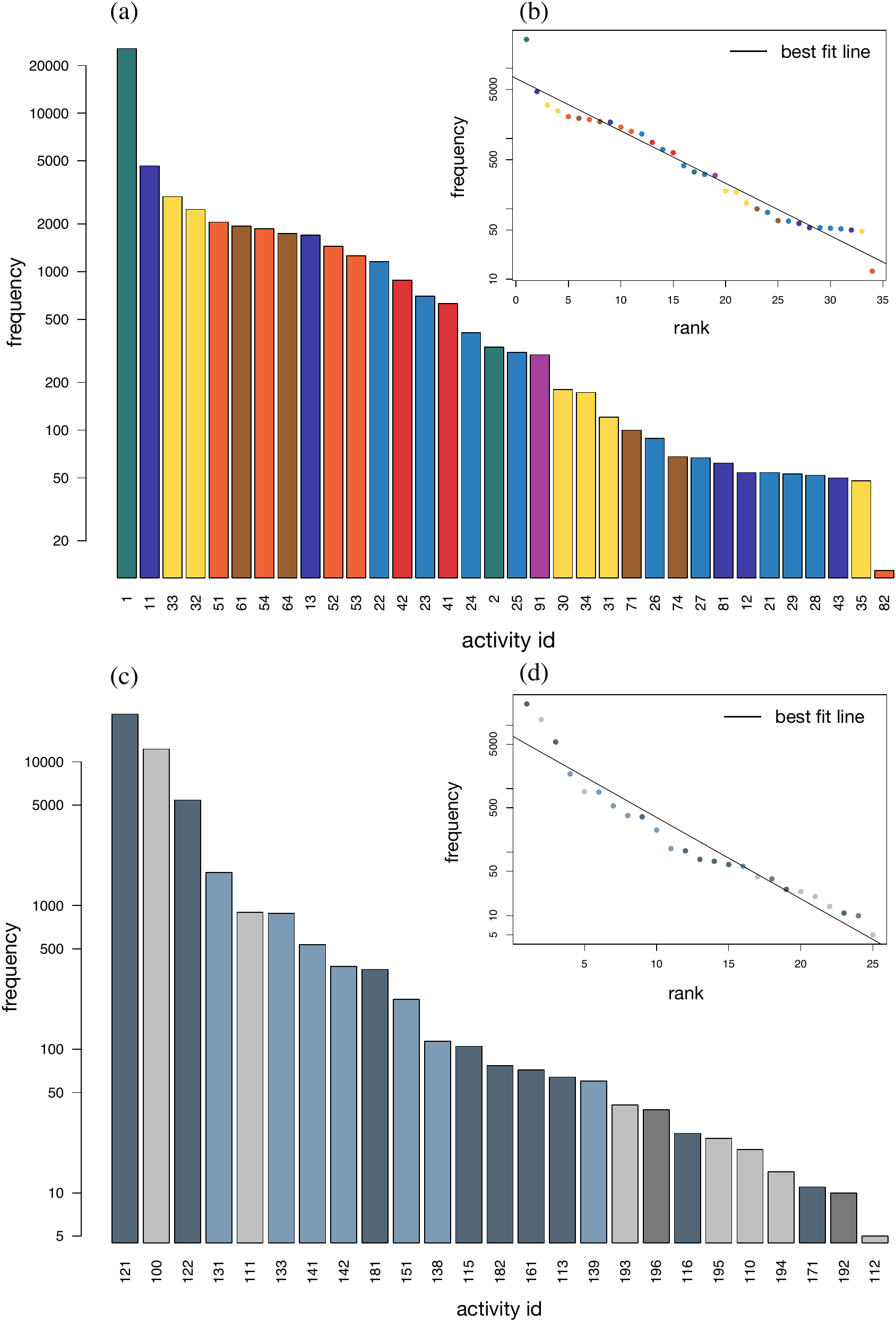}
    \caption{Stop (a) and move (c) activity distribution plot (a) log-plot showing the frequency of each stop activity codes, colors refer to aggregated activity. (b) and (d) show compatibility to a Zipf law model, each point correspond to activities in bar plot below.}
    \label{fig:stop_freq}  
\end{figure}

\begin{figure*}[!h]
    \includegraphics[width=\textwidth]{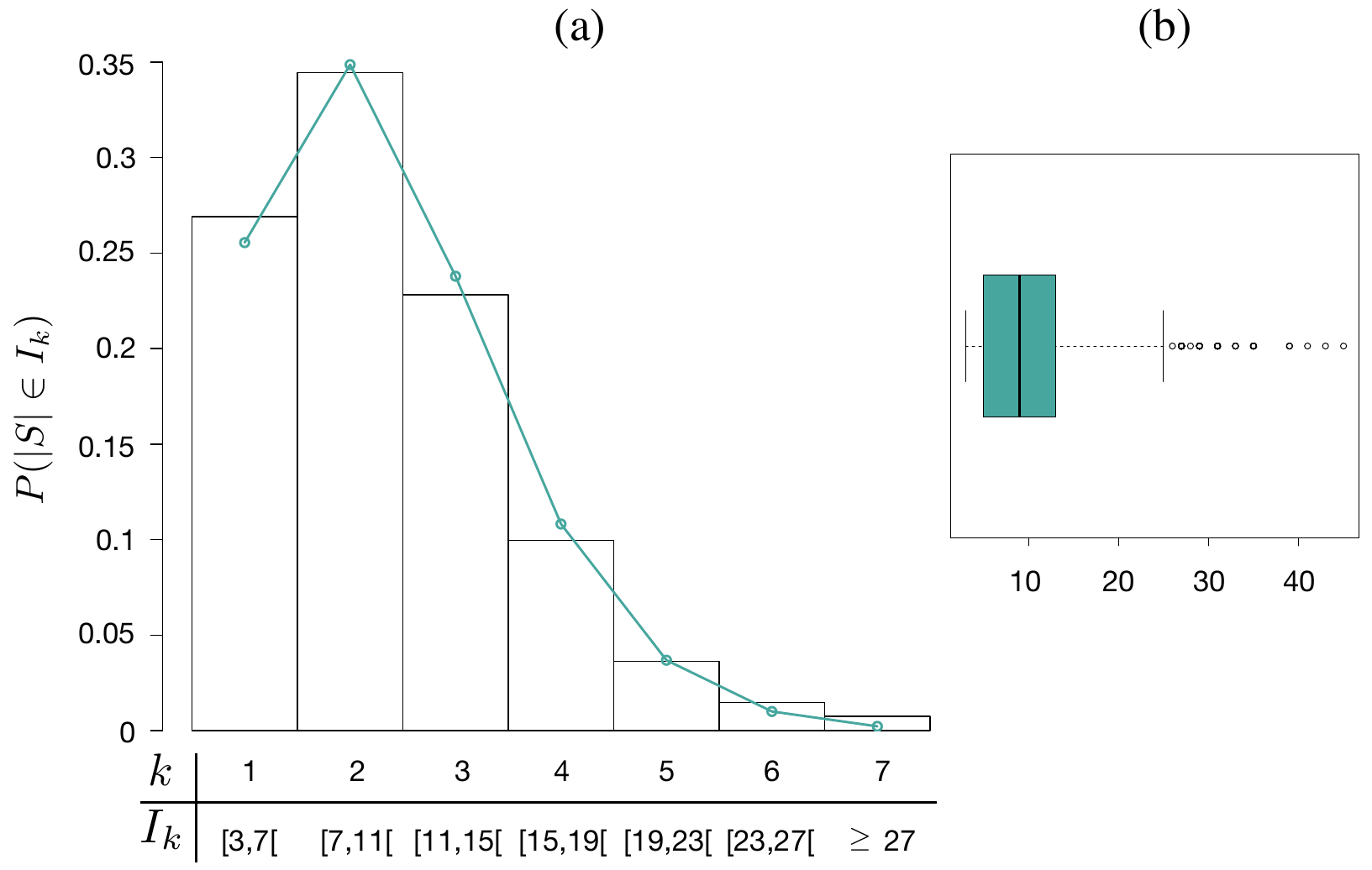}
    \caption{Length statistics of mobility sequences (a) The distribution of length $|S|$ for a given interval $I_{k\in \{1...7\}}$ follows a Poisson distribution $P(|S|\in I_k) \approx \frac{1.36^k e^{-1.36}}{k!}$ (b)  Box plot of the lengths}
    \label{fig:poisson_length}
\end{figure*}

\begin{figure*}[h!]
    \includegraphics[width=.95\textwidth]{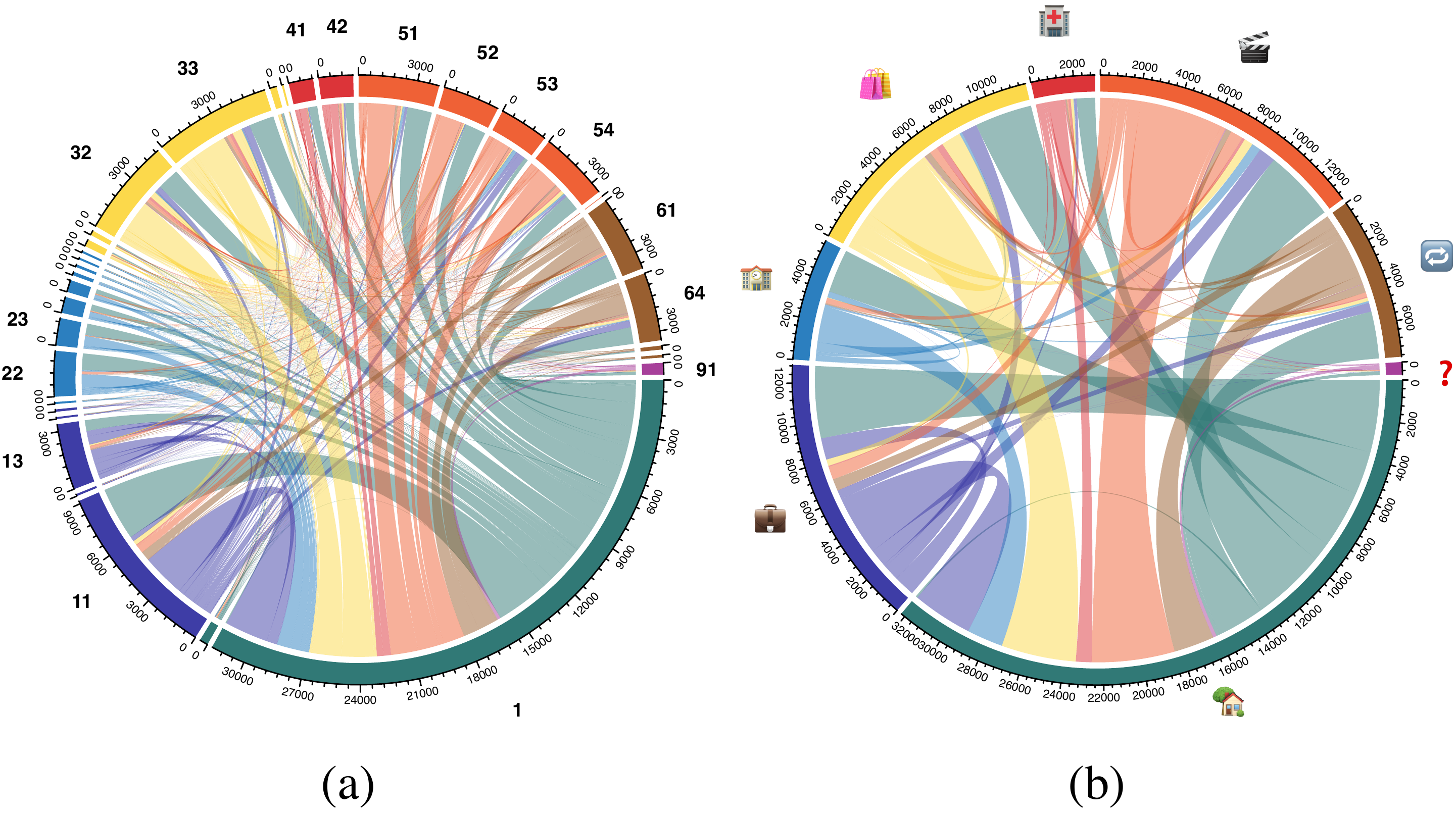}
    \caption{Chord diagram of flows between two consecutive stop activities (a) with all activities (b) with aggregated activities}
    \label{fig:flow}
\end{figure*}

\begin{figure*} [t]
    \includegraphics[width=.9\textwidth]{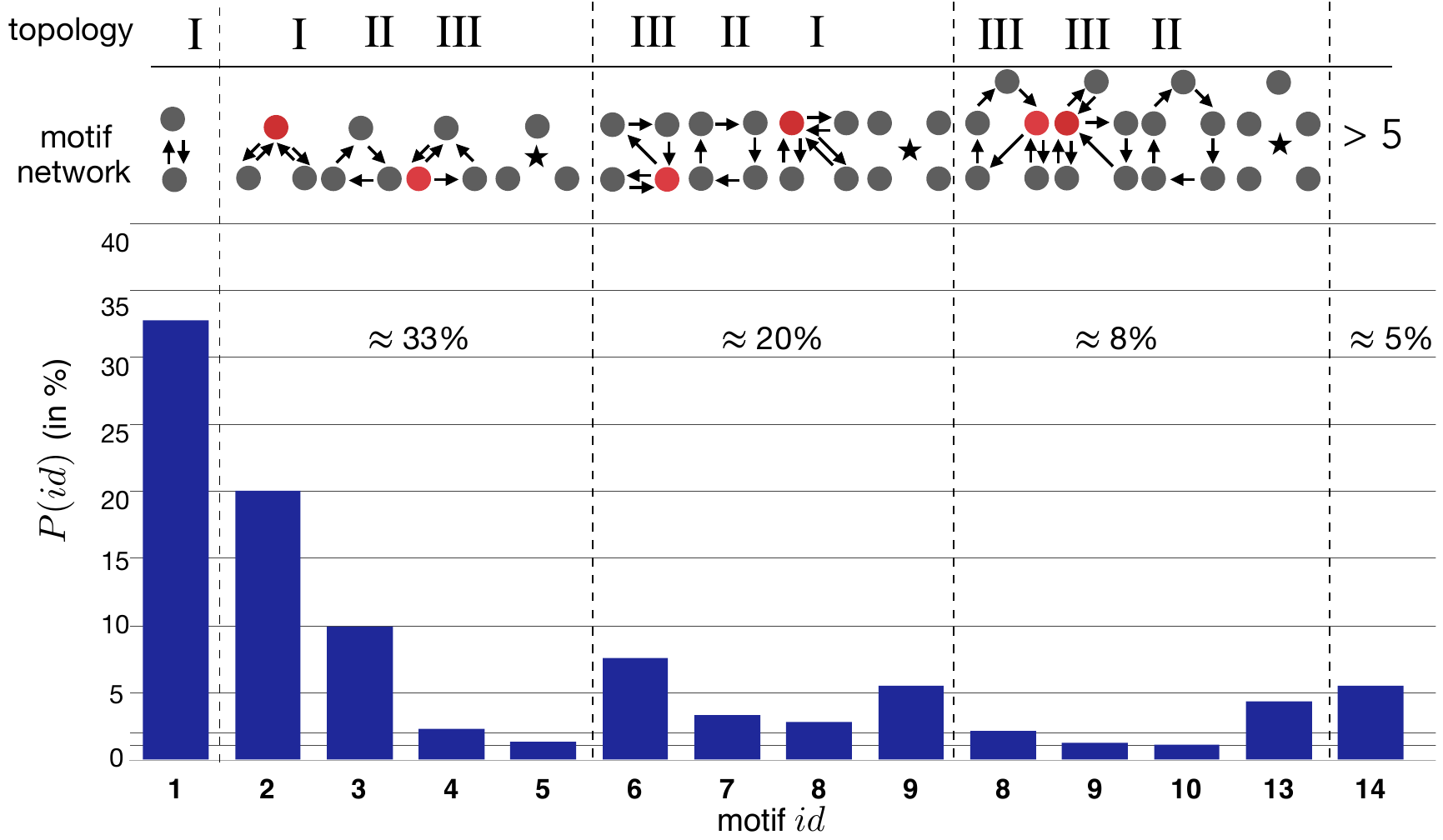}
    \caption{Daily mobility patterns. The motifs are grouped according to their size (separated by dashed lines). $\star$ motifs include all other motifs with $k\in \{3,4,5\}$ nodes. For each group, we show the estimated probability that a given motif has $k$ nodes. The central nodes are highlighted in red. Motifs are classified by three rules indicating topological properties: (I)  graphs with oscillations between two nodes, (II) graphs with cycles of 3 or more nodes and (III) graphs which combining both previous properties (I) and (II).
    }
   \label{fig:daily_patt}
\end{figure*}

\begin{figure*}[h!]
    \includegraphics[width=.9\textwidth]{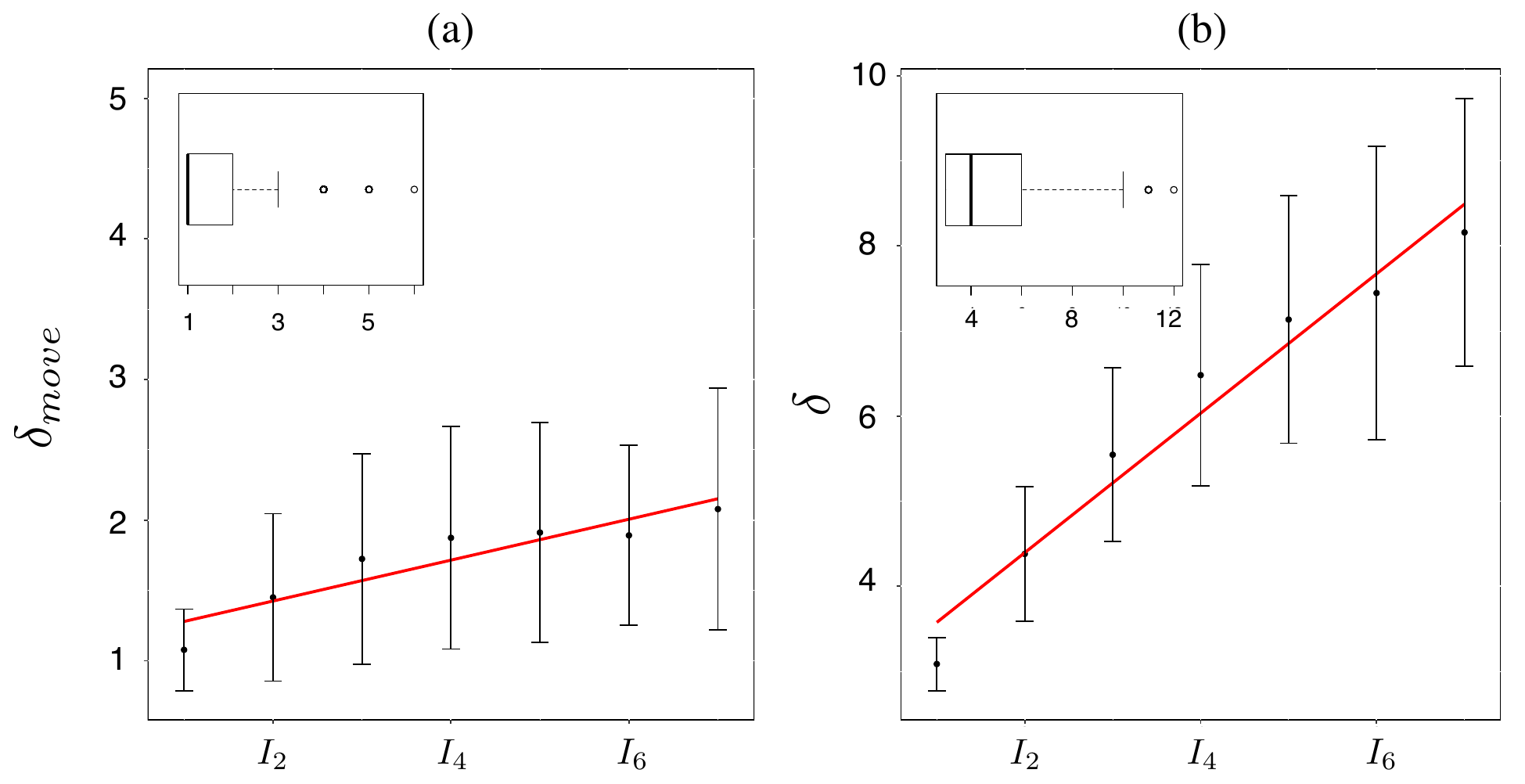}
    \caption{Correlation plots between intervals of length $I_k$ and the number of (a)  distinct move activities $\delta_{move}$, (b) distinct move + stop activities $\delta$ in sequences. Box plot is showed for $\delta$ and $\delta_{move}$. The coefficient of correlation is respectively (a) $\rho = 0.4$ and (b) $\rho = 0.8$.}
    \label{fig:delta}  
\end{figure*}

\begin{figure*}[t]
\centering {
      \includegraphics[width=.98\textwidth]{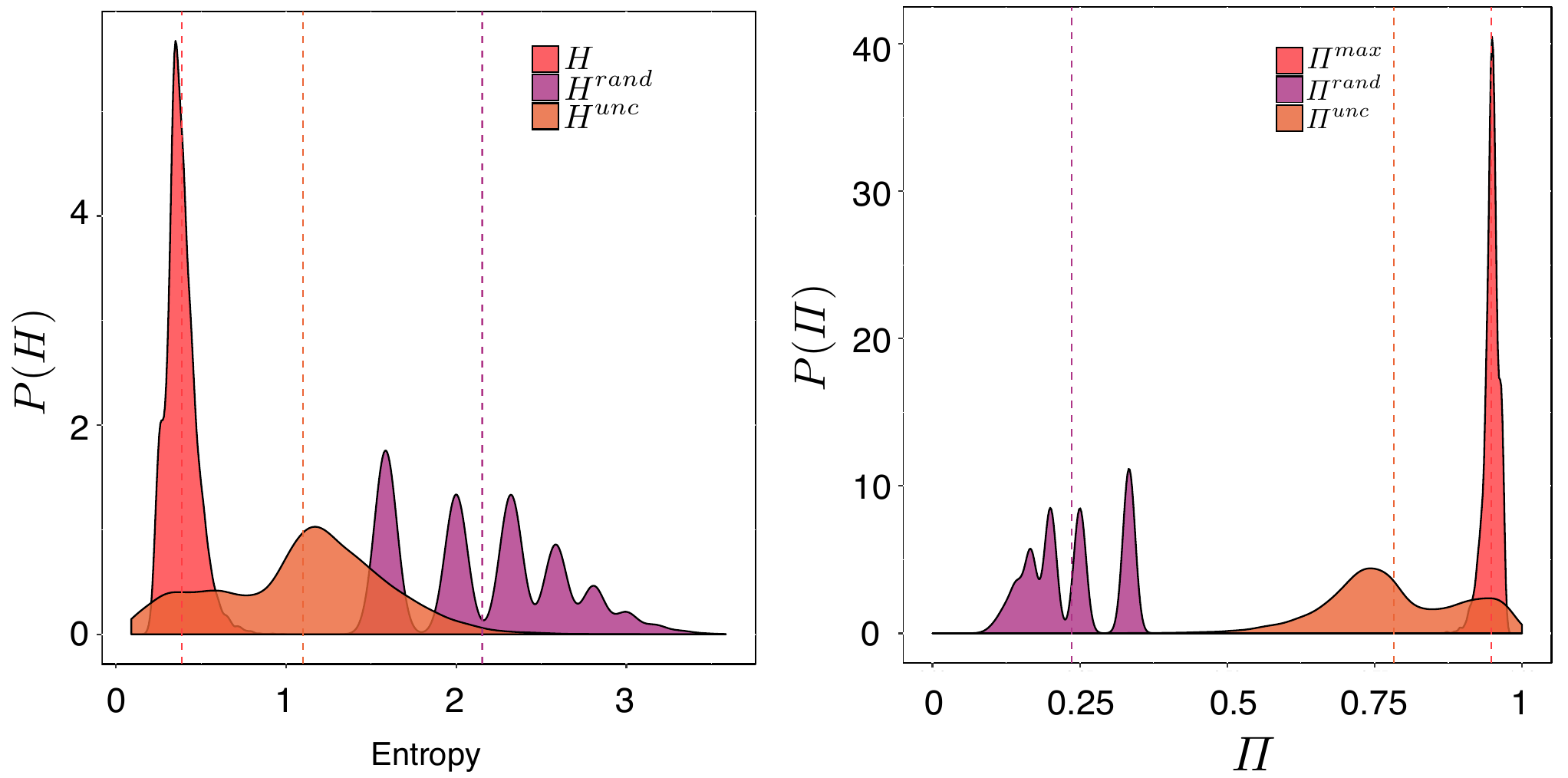}
    }
    \caption{Entropy and predictability of the sequences, dash lines represent the mean (a) probability density function of the entropy $H$,
    the random entropy $H^{rand}$, and the uncorrelated entropy $H^{unc}$ (b) Probability density function of the $\Pi^{max}$, $\Pi^{rand}$, and
    $\Pi^{unc}$}
    \label{fig:entropy}  
\end{figure*}

To understand the meaning of the data, we analyzed the entire dataset using the indicators described in Section \ref{sec:indicator} and summarized in \ref{tab:chosen_indic}. 

Our first elementary study focused on the frequency of each activity in the sequences. For convenience, we separated the stop and move activities. Fig. \ref{fig:stop_freq} presents the distribution of each activity in the dataset. As predicted in \cite{Song10b}, the frequency distribution follows a Zipf law. Intuitively, the three most frequent stop activities are 1 (home), 10 (work), and 33 (shopping in medium and small shops). For move activities, the most frequent items are 121 (car driving), 100 (walking), and 122 (car riding). This figure highlights the main activities that comprise the sequences.

We also performed a complementary study on the number of activities performed per day by an individual. Based on the stop-move representation, there are very few even sequence lengths. To overcome this issue, we consider intervals of length $I_k$. Fig. \ref{fig:poisson_length} presents the distribution of the lengths of the mobility sequences in the dataset. The green curve represents the estimated probability mass function of a Poisson distribution with a parameter $\lambda$ obtained from maximum-likelihood estimation ($\lambda = 1.36$). One can see that the intervals of lengths fit the Poisson distribution.

Another method for semantic sequence analysis is to study the transitions between symbols using an origin-destination matrix. Fig. \ref{fig:flow} presents the transitions between two consecutive stop activities in the dataset. The ontology allows us to visualize these flows according to different levels of granularity. Detailed activities are presented on the left and aggregated activities are presented on the right. One can see that the home activity (\emoji{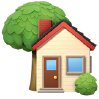}{9}) plays a major role for most transitions, where \emoji{Figures/Emoji/home.pdf}{9} $\rightarrow x$ and the reverse $x \rightarrow$ \emoji{Figures/Emoji/home.pdf}{9}.

In the daily mobility context, transitions were also studied in terms of individual mobility networks to identify topological patterns. Based on the work by Schneider et al. \cite{Schneider13}, we extracted the main motifs from the sequences. As shown in Fig. \ref{fig:daily_patt}, the extracted motifs and frequencies are consistent with the results presented in \cite{Schneider13}. We show the three most frequent motifs for $3, 4$ and $5$ nodes. 
We present the three most-frequent motifs for groups of three, four, and five nodes. Globally, one can see that the most-frequent patterns have less than four nodes and exhibit oscillations (labels I and III). Approximately 87\% of the sequences follow one of the 11 identified motifs. Additionally, this analysis demonstrates that mobility sequences contain many stop activity repetitions. 

Another technique for studying the repetition and regularity of a sequence $S$ is to calculate the  number of unique symbols $\delta$ it contains. Fig. \ref{fig:delta} presents the correlation between the length of a sequence $|S|$ and the distinct number of activities $\delta$. The horizontal axis represents the interval length defined in Fig.  \ref{fig:poisson_length} and the vertical axis represents the numbers of distinct moves $\delta_{move}$  (left side) and number of distinct activities (stops + moves) $\delta $ (right side). One can see that $\delta_{move}$ remains globally stable with one or two different modes for any length of sequence. Therefore, we know that the diversity in the sequences stems from stop activities, while move activities are more often repeated. Regardless, according to the red curve in \ref{fig:delta}(b), one can see that most activities are repeated in a sequence.

Finally, the entropy and predictability of the mobility sequences can be studied to determine how sequences can be predicted. Fig. \ref{fig:entropy} portrays the distributions of these two variables. According to the number of activities in the sequences, the results are similar to those given by \cite{Song10b} and exhibit a low real uncertainty regarding a typical individual’s location $2^{0.4}\approx 1.32$, which is less than two activities). 
It should be noted that these results are consistent with those presented Fig. \ref{fig:delta} for the $\delta$ values. 
The predictability in the random case is $\Pi^{rand}\approx 0.24$. One can see that the median number of different concepts in a sequence is four. This means that we can typically predict one out of the four previous activities. Unlike Song et al.’s results, our $P(\Pi^{unc})$ distribution peaks approximately at $\Pi^{unc} \approx 0.78$ which is similarly to the $\Pi^{rand}$ value. This finding can be explained first by the small number of distinct activities in the sequences and also by the relatively samll number of concepts in the dataset and the Zipf laws they follow (Fig. \ref{fig:stop_freq}). This allow us to predict certain key activities (e.g., home, car, working, walking) based on the user activity history. 
Finally, the real predictability $P(\Pi)$ is peaked near $\Pi^{max} \approx 0.95$, indicating that having a historical record of the mobility of an individual yields a high degree of predictability.

\section{Semantic clustering behavior}
\label{sec:sem_clust}

This section describes the application of the steps (c) and (d) presented in the pipeline Fig. \ref{fig:overview} applied to the EMD Rennes 2018 dataset. The first subsection describes the clustering process using the CED similarity measure and the HAC clustering algorithm Agnes \cite{Kaufman09} with R software. We cluster the individual semantic mobility sequences and analyse variations in daily activity types. In the second subsection, we extract typical behaviors from clusters by summarizing main characteristics and distinct patterns in terms of the indicators discussed in Sections \ref{sec:indicator} and \ref{sec:clust_anal_method}. A discussion of the obtained results and alternatives methods concludes this section. 

\begin{figure*}[t]
    \includegraphics[width=\textwidth]{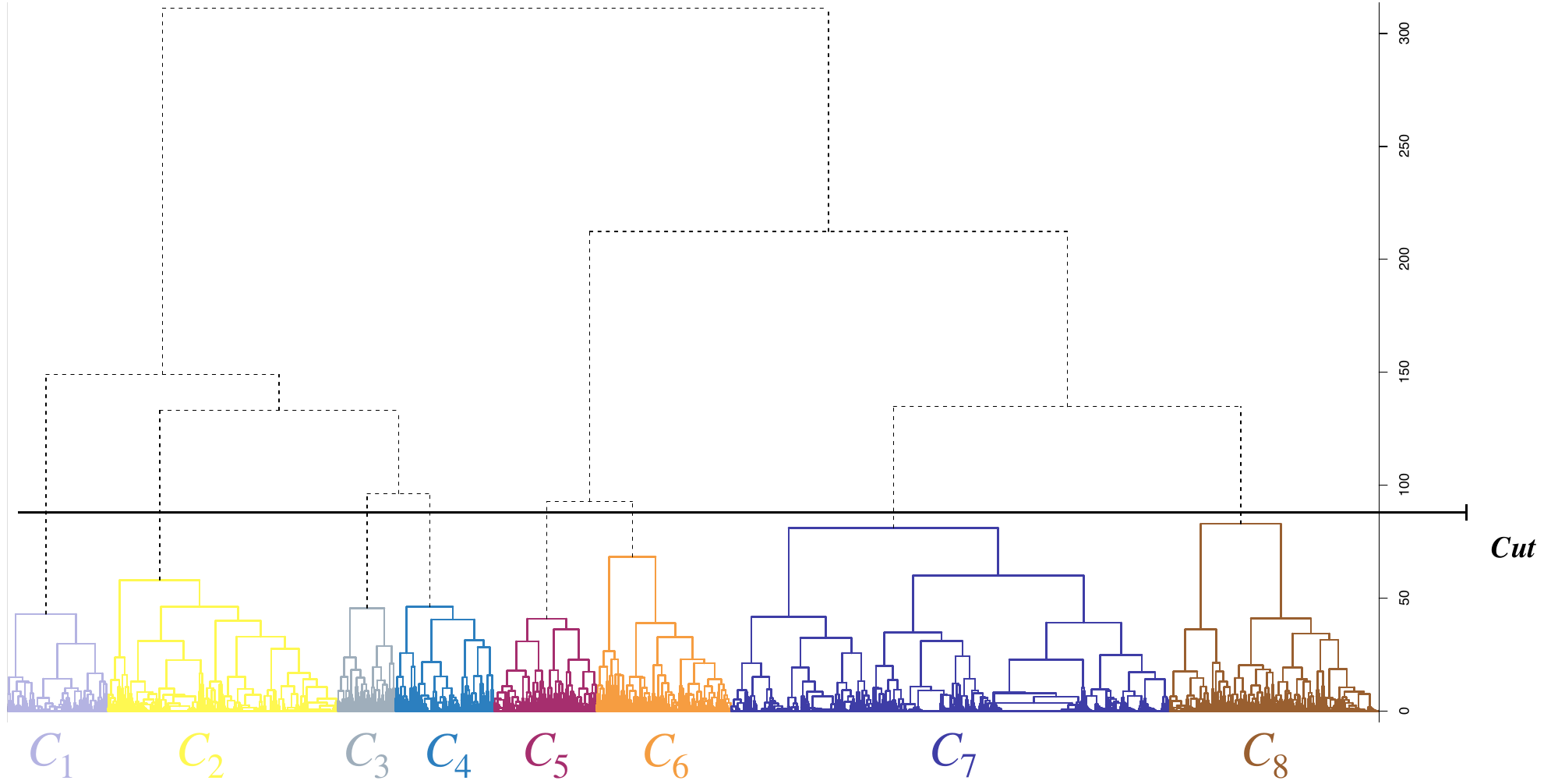}
    \caption{Dendrogram of the HAC clustering algorithm of the EMD 2018 dataset. Eight clusters are formed by the cut of the dendrogram.}
    \label{fig:dendro}  
\end{figure*}

%%%%%%%%%%%%%%%
\subsection{Clustering process}
%%%%%%%%%%%%%%%

As discussed in Section \ref{sec:clust_anal_method}, the clustering process is performed based on the CED measure and a hierarchical clustering algorithm. Here, we discuss the settings for these two methods and the validity of the clusters obtained in terms of quality scores. 

%%%%%%%%
\subsubsection{Similarity measure and HAC initialisation}
%%%%%%%%

As described in Section \ref{sec:ced}, the CED similarity measure requires the setting of several parameters. Empirically, we applied the following settings for CED during the clustering process:
\begin{itemize}
    \item The $\alpha$ coefficient is set to zero to give fully priority to context.
  \item The contextual vector was encoded using the Gaussian kernel bellow. 
  \[f_k(i)=\exp \left( -\frac{1}{2} \left( \frac{i-k}{\sigma} \right)^2 \right) \] 
  where $\sigma$ is a coefficient that controls the flatness of the curve around the activity at position $k$. The larger is $\sigma$, the more context surrounding the index $k$ is considered. In our experiments, we used the value of $\sigma = \frac{m}{2}$ where $m$ is the median sequence size ($m=9$ according to Fig.  \ref{fig:poisson_length}). Therefore, $v_i(e) = f_k(i)$.
\end{itemize}

With these settings, the CED is a semi-metric, meaning it satisfies the requirements of symmetry and identity of indiscernible, but the triangle inequality does not hold. 

Regarding the HAC algorithm, because we do not know the shapes of the clusters, but we want to preserve robustness to outliers and immunity to chain effects, we propose using the Ward criteria, which minimizes the total within-cluster variance, leading to the generation of convex compact clusters that are less affected by noise. 

%%%%%%%%
\subsubsection{Clustering validity}
%%%%%%%%
\label{sec:cluster}

One problem in an unsupervised clustering process is to determine the optimal number of clusters that best fits the inherent partitioning of the dataset. In other words, we must evaluate the clustering results for different cluster numbers, which is the main problem in determining cluster validity \cite{Halkidi01}. There are three main approaches to validating clustering results: (1) external criteria, (2) internal criteria, and (3) relative criteria. Various indices are available for each criterion. 

The structure of HAC and the formed clusters are presented in Fig. \ref{fig:dendro}. In our study, because we did not have a predetermined cluster structure, we used internal validation indices, whose fundamental goal is to search for clusters whose members are close to each other and far from the members of other clusters. Specifically, we used two indices to select the optimal number of clusters. The first is the \textit{inertia gap}, which represents the total distance between two consecutive agglomerations. The wider is the gap, the greater the change in cluster structure and the greater the Silhouette index, which reflects the compactness and separation of clusters. The \textit{Silhouette index} is defined in the range $[-1,1]$. A higher value Silhouette index indicates a better clustering result. 

\begin{table*}
\caption{Cardinal number, Silhouette indice and diameter and radius of each cluster}
\label{tab:clusters} 
{}%
\begin{tabular}{m{.9cm}cm{1cm}ccccc}
\hline 
\centering{\footnotesize{Cluster $C_{i}$}} & {$|C_{i}|$} & \% (in total) &  {$Sil(C_{i})$} & {$diam(C_{i})$} & {$diam(C_{i}^{95\%})$} & {$rad(C_{i})$} &
{$rad(C_{i}^{95\%})$}\tabularnewline
\hline 
\centering{\footnotesize{1}} & {738} & \centering{{\footnotesize{7.4}}} & {0.41} & {8.85} & {\footnotesize{5}} & 5.04 & 3.44
\tabularnewline
\centering{\footnotesize{2}} & {1673} & \centering{\footnotesize{16.7}} & {0.37} & {20.03} & {\footnotesize{8}} & 12.66 & 4.68
\tabularnewline
\centering{\footnotesize{3}} & {423} &
\centering{\footnotesize{4.2}} & {0.01} & {20.81} & {\footnotesize{7.74}} & 7.95 & 5.53
\tabularnewline
\centering{\footnotesize{4}} & {719} &
\centering{\footnotesize{7.2}} & {0.12} & {26.64} & {\footnotesize{7.21}} & 12.51& 5.7
\tabularnewline
\centering{\footnotesize{5}} & {747} &
\centering{\footnotesize{7.5}} &{0.18} & {23.42} & {\footnotesize{6.9}} & 9.86& 5.35
\tabularnewline
\centering{\footnotesize{8}} & {981} &
\centering{\footnotesize{9.8}} &{0.1} & {24.34} & {\footnotesize{8.15}} & 14.59& 6.11
\tabularnewline
\centering{\footnotesize{7}} & {3199} &
\centering{\footnotesize{32}} & {0.29} & {20} & {\footnotesize{5.57}} & 11.09& 4.09
\tabularnewline
\centering{\footnotesize{12}} & {1525} &\centering{\footnotesize{15.2}} & {0.07} & {28.5} & {\footnotesize{7}} & 10.14 & 4.65
\tabularnewline
\hline 
\end{tabular}{\footnotesize \par}
\end{table*}

\begin{figure*}[t]
    \includegraphics[width=\textwidth]{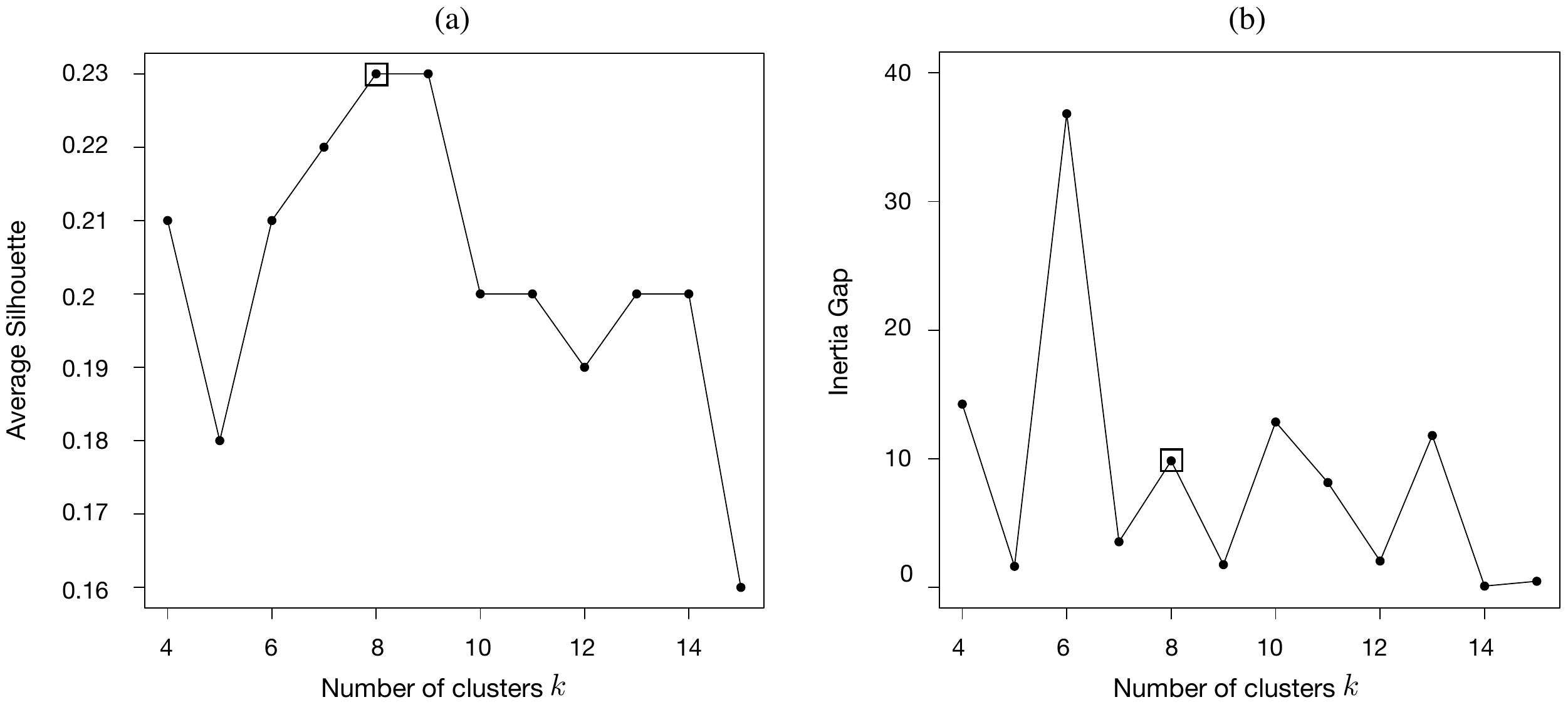}
    \caption{Clustering validity indices (a) Average Silhouette (b) Inertia gap}
    \label{fig:silhouette}  
\end{figure*}

Fig. \ref{fig:silhouette} presents graphs of (a) the average Silhouette index and (b) inertia gap. The relatively low Silhouette values can be attributed to the particular topology associated with CED combined with the Ward criterion and the presence of outliers.\footnote{Note that Silhouette is particularly suitable for hyper-spherical clusters like the one constructed by K-means algorithms.}.
Because we want a number of clusters greater than five to ensure correct analysis, plot (a) suggests the choice of eight or nine clusters. Values of 6, 7, 10, 11, 13, or 14 could also be used. Plot (b) strongly encourages the choice of six clusters, but 8, 10, or 13 clusters could also be used. According to these results, we \textbf{set the number of clusters to eight} for further analysis. Regardless, the choice of six clusters for narrow analysis, or 10 or 13 clusters for wide analysis may be feasible. 

Additional information regarding the clusters, such as proportions, Silhouette index, diameters and radii are given in Table \ref{tab:clusters}. $C_i^{95\%}$ indicates that we filtered 5\% of the most extreme values from the distribution. Therefore, the difference between $diam(C_i)$ and the diameter of 95\% of the elements in $C_i$, denoted as $C_i^{95\%}$, indicates that there is a proportion of outliers far away from the other elements in the cluster $C_i$. Similar radii values of $C_i^{95\%}$ support this analysis.

%%%%%%%%%%%%%%%
\subsection{Behavior extraction and cluster explanation}
%%%%%%%%%%%%%%%
\label{sec:behavior_extract}

\begin{figure*}[t]
    \includegraphics[width=.9\textwidth]{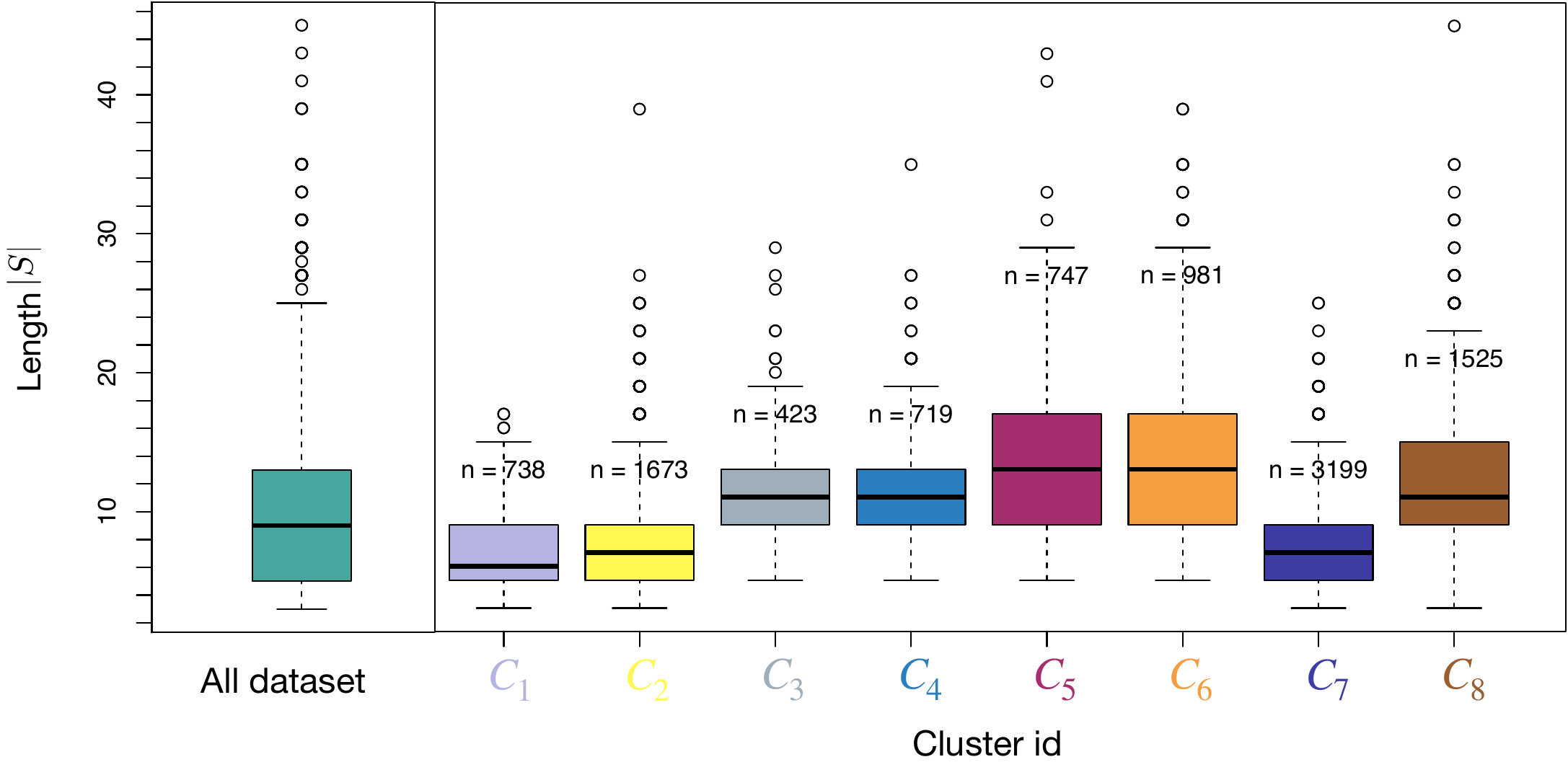}
    \caption{Box plots of sequences' length in each cluster}
    \label{fig:size_clust}  
\end{figure*}

In this section, we reuse the indicators and statistics presented in Table \ref{tab:chosen_indic} and Section \ref{sec:case_study}, but enhanced with significance tests, to infer typical behaviors from clusters discovered in Section  \ref{sec:cluster}. This can help us to check the validity and interpretability of their patterns. 

\begin{figure*}
    \includegraphics[width=.8\textwidth]{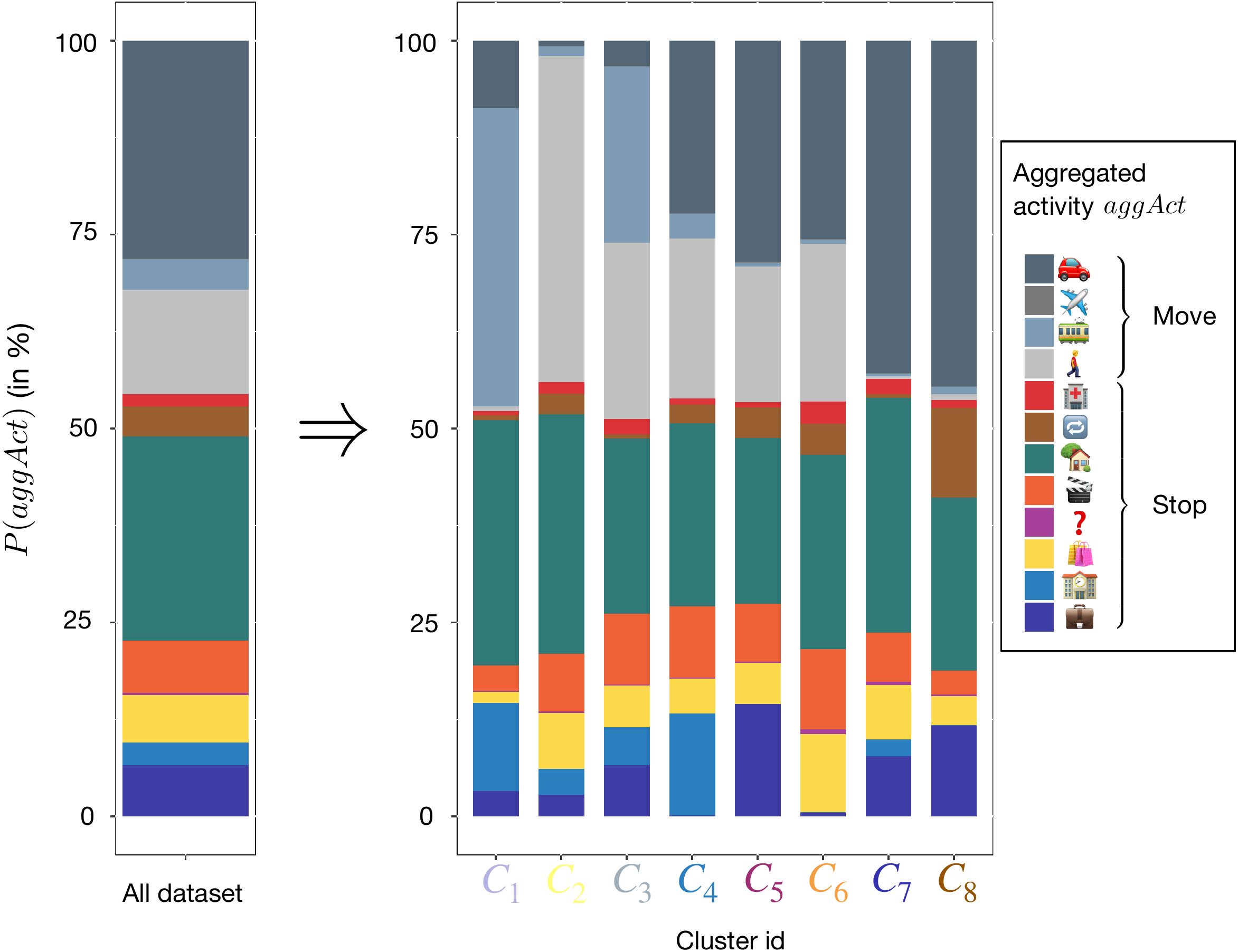}
    \caption{Stacked plot of the proportion of aggregated activities ($aggAct$) in all dataset on the left and in each cluster on the right}
    \label{fig:stack}  
\end{figure*}

First, we analyse the lengths of the sequences inside the clusters. Fig. \ref{fig:size_clust} presents the box plots for the sequence lengths in each cluster. Compared to the distribution of lengths and the box plot for the entire dataset (leftmost plot), one can see that clusters $C_1, C2$ and $C_7$ contain relatively short sequences with a median lengths of six and seven activities, corresponding to intervals $I_1$ and $I_2$ in the Poisson distribution (see Fig. \ref{fig:poisson_length}). In contrast, clusters $C_5, C_6$ and $C_8$ contain longer mobility sequences but have large length dispersions. Analogously, clusters $C_3$ and $C_4$ have middling sequence lengths corresponding to intervals $I_2$ and $I_3$. 
The overlapping of box plots (i.e., the existence of several clusters containing sequences of the same length) and the distribution of outliers in the clusters indicate that sequence length as not a major criteria for grouping sequences. We claim that is an advantage of CED w.r.t other OM similarity measures.

Regarding the distributions of activities, Fig. \ref{fig:stack} portrays the proportions of aggregated activities\footnote{Thanks to the ontology, we can select the level of granularity of our analysis. Aggregated activities having been retained in order to avoid cognitive overload on graphs.}.
An interesting effect that can be observed in this graph is the strong discrimination and stratification effect of clusters according to move activities. Motorized transport activities are very common in clusters $C_7$ and $C_8$, but other move activities are not. Clusters $C_2$ to $C_6$ stand out based on their large proportion of smooth move activities whereas $C_1$ and $C_3$ contain many public transportation move activities. 
Several stop activities are also a distinctive features in certain clusters. For example, school activities are particularly popular in clusters $C_1$ and $C_4$, while work activities in clusters $C_5, C_7$ and $C_8$ and accompany activities being common in cluster $C_8$. Similarly, some clusters tend to contain very few instances of certain activities. For example, cluster $C_6$ contains very few work or study activities

\begin{figure*}
    \includegraphics[width=.65\textwidth]{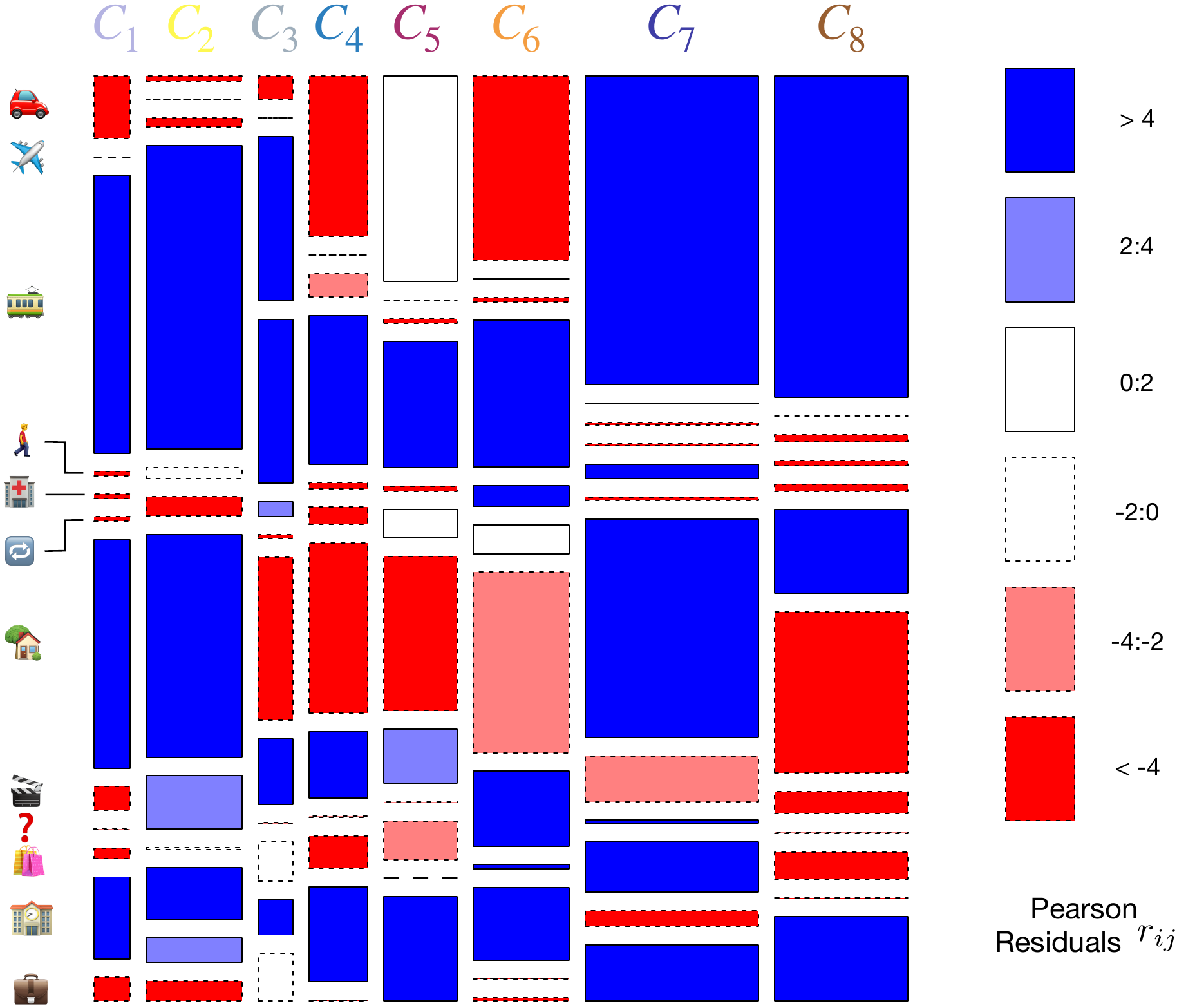}
    \caption{Mosaic plot and Pearson residuals between aggregated activities and clusters. Cramér's $V = 0.3$}
    \label{fig:mosaic}  
\end{figure*}

To analyze theses over- and under-representations, we generated a mosaic plot combined with Pearson residuals to quantify the departure of each cell from independence. Fig. \ref{fig:mosaic} presents the mosaic plot with residuals between clusters and aggregated activities.

We now recall several rules for the interpretation of this type of plot. Let each line represent a cluster an aggregated activity $aggAct_i$ and each column represents $C_j$. We let $c_{ij}$ denote the cell in line $i$ and column $j$:
\begin{itemize}
    \item The width of $c_{ij}$ is proportional of the size of $C_j$.
    %: $width(c_{ij}) \propto n_{+j}$
     \item The height of $c_{ij}$ is proportional the number of $aggAct_i$ under the condition of to be in $C_j$.
     %: $height(c_{ij}) \propto \frac{n_{ij}}{n_{+j}}$
     \item The area of $c_{ij}$ is proportional to the frequency of $aggAct_i$ and $C_j$.
     %: $area(c_{ij}) \propto n_{ij}$.
\end{itemize}

The color of a cell $c_{ij}$ indicates the value of the corresponding Pearson residual $r_{ij}$. A blue shaded cell indicates an over-representation of the aggregated activity $aggAct_i$ in $C_j$. A red-shaded cell indicates an under-representation.
Based on this graphical representation, it is easy to visualize the proportion of a given activity in each cluster. For example, one can immediately see that approximately 40\% of cluster $C_1$ is comprised of public transportation activities. Additionally, based on the residuals, we can immediately and easily identify the characteristic activities in a cluster, as well as those that are under-represented Therefore, the Fig. \ref{fig:mosaic} complements and validates our previous analysis based on a stacked plot (Fig. \ref{fig:stack}) based on quantitative Pearson residuals. The Cramér's $V$ coefficient provides information regarding the associations between clusters and aggregated activities. The good value of $V$ (0.3) highlights the quality of association between our clusters and the activities performed in mobility sequences. The low number of white cells in the mosaic plot confirms our choice of clustering process. 

\begin{landscape}

\begin{figure*}[p]
\includegraphics[width=1.4\textwidth]{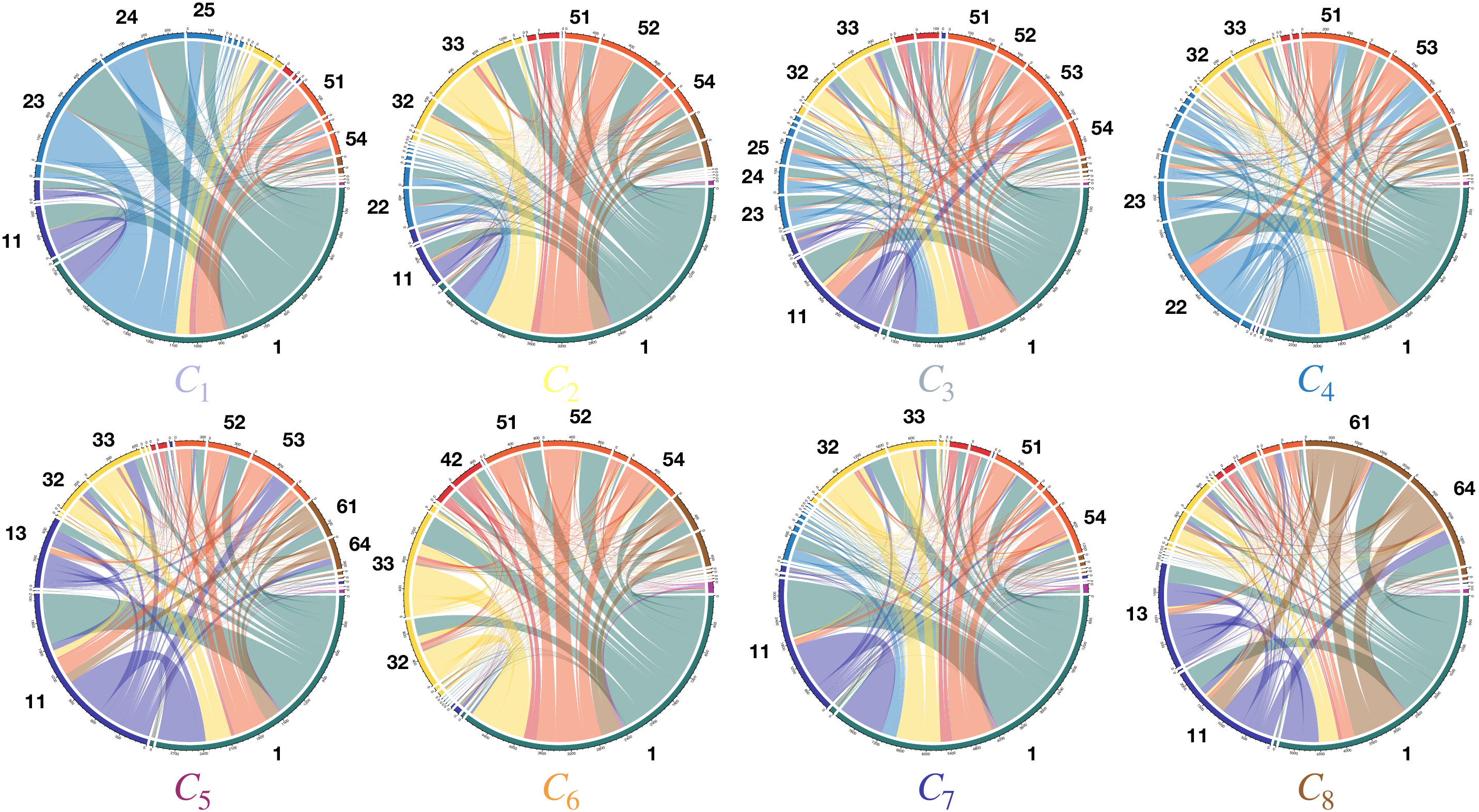}
\caption{Chord diagrams of Stop activities in each cluster}
\label{fig:flow_clust}  
\end{figure*}
\end{landscape}

Regarding transitions between activities, Fig.  \ref{fig:flow_clust} presents a chord diagram for each cluster. As a consequence of the analysis presented in Fig. \ref{fig:delta} which indicated transport mode remains globally stable within sequences, we only represent stop activity transitions. Flows are represented between two leaf activities in the ontology to explore the content of clusters in detail. For example, one can see that cluster $C_1$ contains study activities ranging from junior high school (23) to university (25), whereas cluster $C_4$ mainly contains school children (22).
An explanation for this split can be derived from the fact that cluster $C_1$ mainly concentrates on public transportation activities (see Figs. \ref{fig:stack} and \ref{fig:mosaic}), which is generally related to teenagers, whereas older students tend to be autonomous. However, younger children are mainly accompanied to school by their parents by car or on foot, which can be observed in cluster $C_4$. This analysis is supported by Table 8, where the centrality indicators highlight typical sequences. For example, the most frequent sequence in $C_1$ (mode) involves traveling to and from school via public transportation. Cluster $C_4$ mainly focuses on car and foot travel, but also includes some leisure activities (51, 53). 

\begin{figure*}
    \includegraphics[width=\textwidth]{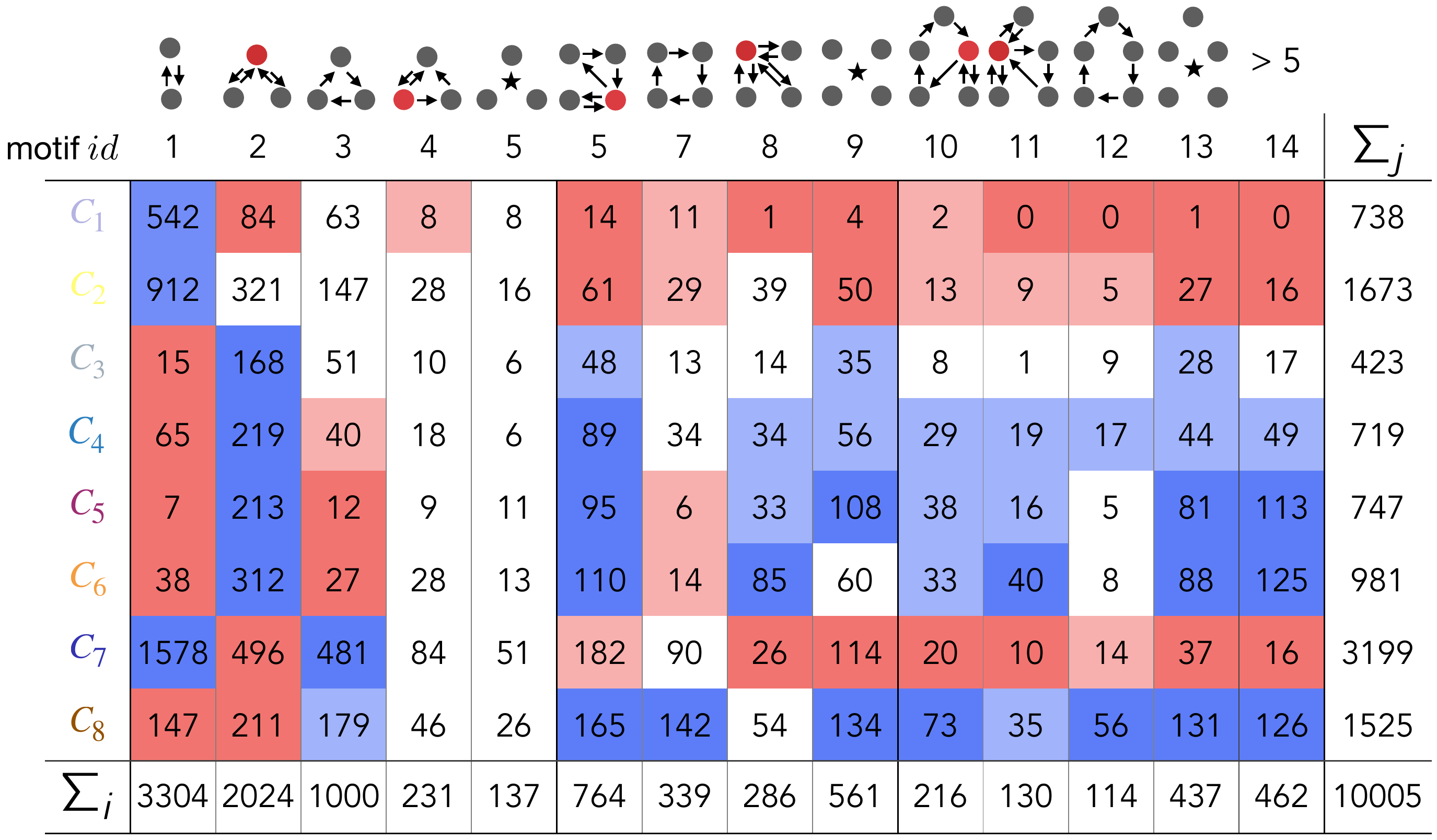}
    \caption{Heat map with Pearson residuals of daily patterns in each cluster. Cramér's $V = 0.25$}
    \label{fig:motif_clust}  
\end{figure*}

Regarding the worker cuters (i.e, $C_5, C_7$ and $C_8$), we observed different behaviors for each one. In cluster $C_5$, the typical behavior appears to be that of a worker driving to work (11, 13) and then walking to a restaurant for lunch (53) before returning to work and then driving home. This scenario is supported by the medoid mobility sequences and Fig. \ref{fig:motif_clust} which represents the daily patterns in each cluster. In $C_5$, one can see a trend of oscillation between two activities with a central node. There are also some activities that can be added to the semantic sequence, such as shopping (32, 33) after work, going for a walk or window-shopping (52), or accompanying activities (61, 64). 
Cluster $C_7$ represents individuals who travel to an activity, typically work (11), by car, then return home by car again. This interpretation is consistent with the short semantic mobility lengths in the cluster and the large majority of daily patterns with a single oscillation. This mobility behavior is the most common in the dataset and can be interpreted as the daily routine of going to work by car, occasionally shopping in a mall (32), and then going back home. Cluster $C_8$ is focused on  workers that accompany and pick up (61, 64) someone before and after work (11, 13) with a possible mobility around the workplace (13). Similarly, cluster $C_7$ moves are almost exclusively performed by car. Furthermore, in this cluster, mobility sequences are relatively long and form complex patterns with, generally, four or more stop activities.

Finally, some clusters can be distinguished by the absence of some common elements. For example, the people in cluster $C_6$ do not work or study and they tend to spend their time mainly on shopping or leisure activities. Additionally, box plot in Fig. \ref{fig:size_clust} reveals that the mobility sequences in $C_6$ are long. Lastly, clusters $C_2$ and $C_3$ are especially characterized by their move activities. Individuals in $C_2$ almost exclusively move on foot to perform a single activity. Compared to the mobility sequences in $C_6$, these sequences are relatively short and based on a single oscillation between home and another location. The typical behavior in $C_3$ seems to be that people who use both public transportation and walking for mobility. 

\begin{table*}[t]
\caption{Centrality indicators in each cluster}
\label{tab:centrality} 
\begin{tabular}{ccc}
\hline
{Cluster $C_{i}$} & {Medoid} & {Mode}
\tabularnewline
\hline
\multirow{2}{*}{{1}} & $\langle$\emoji{Figures/Emoji/home}{12}, \emoji{Figures/Emoji/public}{12}, \emoji{Figures/Emoji/study}{12}, \emoji{Figures/Emoji/motor}{12}, \emoji{Figures/Emoji/home}{12}$\rangle$ & $\langle$\emoji{Figures/Emoji/home}{12}, \emoji{Figures/Emoji/public}{12}, \emoji{Figures/Emoji/study}{12}, \emoji{Figures/Emoji/public}{12}, \emoji{Figures/Emoji/home}{12}$\rangle$ \tabularnewline
 & {$\tuple{1,131,23,122,1}$} & {$\tuple{1,141,23,141,1}$}
 
 \tabularnewline
 
\multirow{2}{*}{{2}} & $\langle$\emoji{Figures/Emoji/home}{12}, \emoji{Figures/Emoji/smooth}{9}, \emoji{Figures/Emoji/shop}{12}, \emoji{Figures/Emoji/smooth}{9}, \emoji{Figures/Emoji/home}{12}$\rangle$ &
$\langle$\emoji{Figures/Emoji/home}{12}, \emoji{Figures/Emoji/smooth}{9}, \emoji{Figures/Emoji/shop}{12}, \emoji{Figures/Emoji/smooth}{9}, \emoji{Figures/Emoji/home}{12}$\rangle$

\tabularnewline

 & {$\tuple{1,100,33,100,1}$} & {$\tuple{1,100,33,100,1}$}\tabularnewline
\multirow{2}{*}{{3}} & $\langle$\emoji{Figures/Emoji/home}{12}, \emoji{Figures/Emoji/public}{12},
\emoji{Figures/Emoji/public}{12},
\emoji{Figures/Emoji/work}{12}, \emoji{Figures/Emoji/smooth}{9},
\emoji{Figures/Emoji/leisure}{12},
\emoji{Figures/Emoji/public}{12},
\emoji{Figures/Emoji/home}{12}$\rangle$  &
$\langle$\emoji{Figures/Emoji/home}{12}, \emoji{Figures/Emoji/public}{12}, \emoji{Figures/Emoji/study}{12}, \emoji{Figures/Emoji/smooth}{9},
\emoji{Figures/Emoji/study}{12},
\emoji{Figures/Emoji/smooth}{9},
\emoji{Figures/Emoji/study}{12},
\emoji{Figures/Emoji/public}{12},
\emoji{Figures/Emoji/home}{12}$\rangle$

\tabularnewline

 & {$\tuple{1,131,131,11,100,53,131,1}$} & {$\tuple{1,141,23,100,27,100,23,141,1}$}\tabularnewline
\multirow{2}{*}{{4}} & $\langle$\emoji{Figures/Emoji/home}{12}, \emoji{Figures/Emoji/motor}{12}, \emoji{Figures/Emoji/study}{12}, \emoji{Figures/Emoji/smooth}{9},
\emoji{Figures/Emoji/leisure}{12},
\emoji{Figures/Emoji/motor}{12},
\emoji{Figures/Emoji/home}{12}$\rangle$ &
$\langle$\emoji{Figures/Emoji/home}{12}, \emoji{Figures/Emoji/motor}{12}, \emoji{Figures/Emoji/study}{12}, \emoji{Figures/Emoji/smooth}{9},
\emoji{Figures/Emoji/home}{12}$\rangle$

\tabularnewline

 & {$\tuple{1,122,22,100,51,122,1}$} & {$\tuple{1,122,22,100,1}$}\tabularnewline
\multirow{2}{*}{{5}} & $\langle$\emoji{Figures/Emoji/home}{12}, \emoji{Figures/Emoji/motor}{12},
\emoji{Figures/Emoji/work}{12},
\emoji{Figures/Emoji/smooth}{9}, 
\emoji{Figures/Emoji/leisure}{12}, 
\emoji{Figures/Emoji/smooth}{9}, 
\emoji{Figures/Emoji/work}{12},
\emoji{Figures/Emoji/motor}{12},
\emoji{Figures/Emoji/home}{12}$\rangle$ &
$\langle$\emoji{Figures/Emoji/home}{12},
\emoji{Figures/Emoji/motor}{12},
\emoji{Figures/Emoji/work}{12},
\emoji{Figures/Emoji/smooth}{9}, 
\emoji{Figures/Emoji/leisure}{12}, 
\emoji{Figures/Emoji/smooth}{9}, 
\emoji{Figures/Emoji/work}{12},
\emoji{Figures/Emoji/motor}{12},
\emoji{Figures/Emoji/home}{12}$\rangle$

\tabularnewline

 & {$\tuple{1,121,11,100,53,100,11,121,1}$} & {$\tuple{1,121,11,100,53,100,11,121,1}$}\tabularnewline
\multirow{2}{*}{{8}} & $\langle$\emoji{Figures/Emoji/home}{12},
\emoji{Figures/Emoji/smooth}{9},
\emoji{Figures/Emoji/shop}{12},
\emoji{Figures/Emoji/smooth}{9}, 
\emoji{Figures/Emoji/home}{12}, 
\emoji{Figures/Emoji/motor}{12}, 
\emoji{Figures/Emoji/leisure}{12},
\emoji{Figures/Emoji/motor}{12},
\emoji{Figures/Emoji/home}{12}$\rangle$ & $\langle$\emoji{Figures/Emoji/home}{12},
\emoji{Figures/Emoji/motor}{12},
\emoji{Figures/Emoji/shop}{12},
\emoji{Figures/Emoji/motor}{12}, 
\emoji{Figures/Emoji/home}{12}, 
\emoji{Figures/Emoji/smooth}{9}, 
\emoji{Figures/Emoji/leisure}{12},
\emoji{Figures/Emoji/smooth}{9},
\emoji{Figures/Emoji/home}{12}$\rangle$

\tabularnewline

 & {$\tuple{1,100,33,100,1,121,52,121,1}$} & {$\tuple{1,121,33,121,1,100,52,100,1}$}\tabularnewline
\multirow{2}{*}{{7}} & $\langle$\emoji{Figures/Emoji/home}{12},
\emoji{Figures/Emoji/motor}{12},
\emoji{Figures/Emoji/work}{12},
\emoji{Figures/Emoji/motor}{12},
\emoji{Figures/Emoji/home}{12}$\rangle$ & 
$\langle$\emoji{Figures/Emoji/home}{12},
\emoji{Figures/Emoji/motor}{12},
\emoji{Figures/Emoji/work}{12},
\emoji{Figures/Emoji/motor}{12},
\emoji{Figures/Emoji/home}{12}$\rangle$

\tabularnewline

 & {$\tuple{1,121,11,121,1}$} & {$\tuple{1,121,11,121,1}$}\tabularnewline
\multirow{2}{*}{{12}} & $\langle$\emoji{Figures/Emoji/home}{12},
\emoji{Figures/Emoji/motor}{12},
\emoji{Figures/Emoji/commute}{12},
\emoji{Figures/Emoji/motor}{12},
\emoji{Figures/Emoji/work}{12},
\emoji{Figures/Emoji/motor}{12},
\emoji{Figures/Emoji/commute}{12},
\emoji{Figures/Emoji/motor}{12},
\emoji{Figures/Emoji/home}{12}$\rangle$  &
$\langle$\emoji{Figures/Emoji/home}{12},
\emoji{Figures/Emoji/motor}{12},
\emoji{Figures/Emoji/work}{12},
\emoji{Figures/Emoji/motor}{12},
\emoji{Figures/Emoji/home}{12}$\rangle$
\tabularnewline
 & {$\tuple{1,121,61,121,11,121,64,121,1}$} & {$\tuple{1,121,13,121,1}$}\tabularnewline
 \hline
\end{tabular}
\end{table*}

%%%%%%%%%%%%%%%
\subsection{Semantic mobility behavior discovering}
%%%%%%%%%%%%%%%

\begin{table*}[t]
\caption{Summary of discovered behaviors}
\label{tab:behavior} 

{}%
\begin{tabular}{ccccm{1.9cm}|c}
\hline 
\centering{Cluster $C_{i}$} & \centering{\% (in total)} & {Typical activities} & {Length} & \centering{Daily Patterns (motif id)} & \centering{\textbf{\emph{Behavior}}}\tabularnewline
\hline 
\centering{1} & \centering{7.4} &\{\emoji{Figures/Emoji/public}{12}, \emoji{Figures/Emoji/study}{12}\} & {Short} & {1} &\centering{ \textbf{Teenagers}}\tabularnewline
\centering{2} & \centering{16.7} & \{\emoji{Figures/Emoji/smooth}{8}, \emoji{Figures/Emoji/shop}{12}\} & {Short} & {1} &\centering{ \textbf{Foot shoppers}}\tabularnewline
\centering{3} & \centering{4.2} & \{\emoji{Figures/Emoji/public}{12}, \emoji{Figures/Emoji/smooth}{8},
\emoji{Figures/Emoji/study}{12},
\emoji{Figures/Emoji/leisure}{12}\} & {Medium} & {2, 5} &\centering{ \textbf{Mixed transportation}}\tabularnewline
\centering{4} & \centering{7.2} & \{\emoji{Figures/Emoji/smooth}{8}, \emoji{Figures/Emoji/study}{12}, \emoji{Figures/Emoji/leisure}{12}\} & {Medium} & {2, 4} &\centering{ \textbf{Schoolchildren}}\tabularnewline
\centering{5} & \centering{7.5} & \{
\emoji{Figures/Emoji/smooth}{8},
\emoji{Figures/Emoji/work}{12},
\emoji{Figures/Emoji/leisure}{12}\} & {Long} & {2, 5, 9, 13, 14} & \centering{\textbf{Wandering workers}}\tabularnewline
\centering{8} & \centering{9.8} & \{\emoji{Figures/Emoji/smooth}{8},
\emoji{Figures/Emoji/shop}{12},
\emoji{Figures/Emoji/leisure}{12}\} & {Long} & {\footnotesize{2, 5, 8, 11, 13, 14}} & \centering{\textbf{Shopping addicts}}\tabularnewline
\centering{7} & \centering{32} & \{\emoji{Figures/Emoji/motor}{12},
\emoji{Figures/Emoji/work}{12}\} & {Short} & {1, 3} & \centering{\textbf{Daily routine}}\tabularnewline
\centering{12} & \centering{15.2} & \{\emoji{Figures/Emoji/motor}{12}, 
\emoji{Figures/Emoji/work}{12}, 
\emoji{Figures/Emoji/commute}{9}\} & {Medium} & {5, 7, 9, 10, 12, 13, 14} & \centering{\textbf{\footnotesize{Working parents}}}\tabularnewline
\hline 
\end{tabular}{\footnotesize \par}
\end{table*}

Based on our previous analysis of clusters, we extract a global behavior from each cluster. Table \ref{tab:behavior} summarizes the eight discovered behaviors. The columns ``Typical activities", ``Length" and ``Daily patterns" were computed using the Algorithm \ref{alg:behavior} and represent the predominant activities, median lengths of sequences (intervals), and the predominant daily patterns, respectively. For the sake of brevity, typical activities were extracted at the aggregated activities level (using emojis). Finally, the ``Behavior" column contains, mnemonic labels which summarizes the analysis carried out Section \ref{sec:behavior_extract}.  

\begin{algorithm}[b]
\SetAlgoLined
\KwData{Set of clusters $\mathcal{C}=\{C_1, ..., C_k\}$}
\KwResult{Typical activities, Length and Daily patterns}
 %\Comment{Build the daily pattern graph of each sequence $S\in \mathcal{D}$} \\
 \For{$C_i \in \mathcal{C}$}{ 
 
 \LeftComment{medoid$(C_i)$ and mode$(C_i)$ refer to Table \ref{tab:centrality}. $f(x,C_i) = \sum_{S\in C_i}\text{count}(x, S)$ denotes the frequency of activity $x$ in all sequences of $C_i$.}
  $\text{Typical activities}(C_i) = \{x | x \in \Sigma \wedge \text{PearsonResiduals}(f(x, C_i)) \geq 4\ \wedge$ \\
   \hspace{5.5cm} $(x_i \in \text{medoid}(C_i) \vee \text{mode}(C_i))\}$  \\
   
\LeftComment{$I_k$ refers to intervals of Poison distribution in Fig. \ref{fig:poisson_length}} \\
$\text{Length}(C_{i})=\begin{cases}
\text{``Short''} & \text{if } median(\{|S|: S \in C_i\})\in I_{1}\\
\text{``Medium''} & \text{if } median(\{|S|: S \in C_i\})\in I_{2}\\
\text{``Long''} & \text{else}
\end{cases}$ \\

\LeftComment{$\mathcal{G}$ refers to a dictionary of networks from Algorithm. \ref{alg:daily_patt}}. \\

$\text{DailyPatterns}(C_{i})=\{G_S |S \in C_i, \text{PearsonResiduals}(\mathcal{G}[G_S]) \geq 4\}$
 }
 \caption{Behavior Discovery summary}
 \label{alg:behavior}
\end{algorithm}

We can summarize the behaviors in the clusters as follow: 
Cluster $C_1$ contains a majority of short mobility sequences, with only one loop between home and middle/high school or university, and an extensive use of public transportation such as buses. This group mainly consists of \textit{Teenagers} mobility behavior.

Cluster $C_2$ is characterized by people who only walk for shopping, which we call the \textit{Foot shoppers}. 

The main feature in $C_3$ is that the sequences combine walking and public transportation, which we call them \textit{Mixed transportation} people. 
\textit{Schoolchildren} are maily clustered in $C_4$, with a large
proportion of primary school activities, followed by sports or cultural activities. These individuals mainly move by walking or by riding in cars.

In cluster $C_5$, the prototypical behavior is that of an individual working and going out for lunch, typically at restaurant. These \textit{Wandering workers} achieved their mobility by driving between home and work and, walking between work and place for food/leisure. 

The representative behavior of individuals in cluster $C_6$ is that they do not work or study. They spent the majority of their time in shopping or leisure activities. We refer to these individuals as the \textit{Shopping addicts}. 

Cluster $C_7$ is the largest cluster and contains 32\% of the dataset. Individuals in $C_7$ done mainly short mobility sequences that represent people who go to work by car and then travel back home. This behavior, with its elementary activities (car, work, and sometimes shopping at a mall) and oscillation patterns, evokes a simple \textit{Daily routine}.

Finally, $C_8$ represents a similar behavior to that of $C_7$ but individuals typically transport somebody else by
car before working and then pick them back up
after work. This behavior can be interpreted as parents accompanying their children to school in the morning and picking them up in the evening. Therefore, we refer to these individuals as \textit{Working Parents}. 

\begin{figure*}
    \centering{
      \includegraphics[scale=.85]{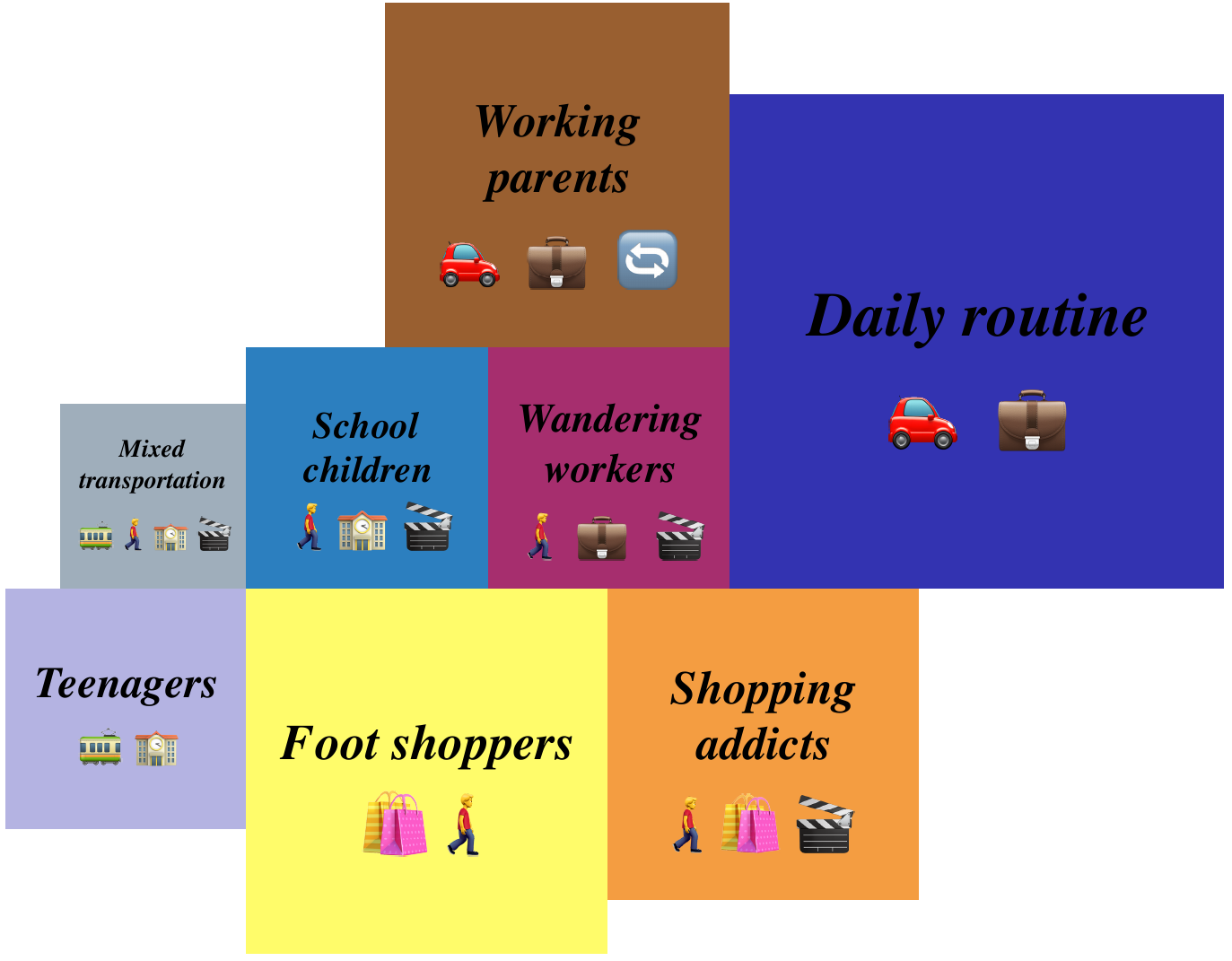}
      }
    \caption{Graphical summary of discovered clusters}
    \label{fig:summary}  
\end{figure*}

Figures \ref{fig:summary} presents a graphical summary of the clusters and corresponding behaviors. The area of each square is proportional to
the size of the associated cluster. The colors and compositions refer to the dendrogram in Figure \ref{fig:dendro}. 

%%%%%%%%%%%%%%%
\subsection{Discussion}
%%%%%%%%%%%%%%%

In the previous subsection, we presented the analysis and results of the clustering process according to the methodology introduced in Section \ref{sec:clust_anal_method}. This facilitated the discovery of several interesting and coherent patterns of semantic mobility, which are summarized in Table \ref{tab:behavior}. 

Regardless, several problems and alternatives should be considered. First of all, as discussed in Section \ref{sec:related_work}, there are many different similarity measures for semantic sequences. The choice of a measure
has a significant impact on the results of clustering. In this paper, we used CED, which is an alternative measure to those mentioned in Table \ref{tab:description}, which could also be 
considered. The setting of CED: the similarity measure between activities, the ontology, the contextual vector and the $\alpha$ coefficient are all parameters that can be modified to change the clustering results. 
We experimentally tuned each parameter and referred to business knowledge for the construction of our ontology. 

The second point is the choice of the clustering algorithm. As indicated in Table \ref{tab:clusters}, the diameters of the clusters indicate the presence of some outliers and the Silhouette scores suggest that the clusters are not hyper-spherical. 
Therefore, the use  of a density-based clustering algorithm such as DBSCAN or OPTICS, combined with more complete study of the topological space and neighborhood relationship via UMAP, as well as intra- and inter-cluster distances, could help us to obtain denser clusters and detect outliers. 

The final point is the level of analysis of the activities in the ontology. To prevent cognitive overload during visualization, Section \ref{sec:behavior_extract} only presented  the results of aggregated activities. Regardless, a detailed analysis at the level of leaf activities in the ontology would also be relevant and could refine the discovered behaviors. 

Based on the proposed methodology, we were able to analyze and extract precise cluster behaviors from both socio-cognitive and urban perspectives. Therefore, this approach should be helpful for expert analysts in terms of limiting psychological biases, such as confirmation bias. 

The proposed methodology supports the comprehension of clusters and is useful for the evaluation and tuning of clustering methods. The discovery of coherent and meaningful behaviors may trigger the proposal of novel metrics for the quality of experimental setups and relevance of various methods.

\section{Conclusions and future work}
\label{sec:conclusion}

In this paper, we introduced a novel methodology, called \textsc{simba}, to mine, discover and analyze behaviors in semantic human mobility sequences. The proposed process is generic and can be adapted for any sequence of categorical data.

\textsc{simba} introduces a simple and complete pipeline from raw data to clustering analysis for studying semantic mobility sequences and extracting  mobility behaviors. \textsc{simba} leverages the use of a hierarchical clustering algorithm combined with CED to cluster similar mobility sequences.

Based on an extended literature review of both human mobility properties and semantic similarity measures, we selected complementary statistical indicators to describe semantic mobility sequences from different points of view. To the best of our knowledge, \textsc{simba} is the first complete and modular methodology supporting the understanding of human 
behaviors with a large panel of visual 
indicators that highlight the complementary properties of semantic mobility. 

The proposed approach was tested on a real dataset of 10\!  005  semantic mobility sequences from a household travel survey. 
We were able to identify specific behaviors that can constitute key information on urban activities. 
Thanks to the proposed methodology, discovered clusters are easily interpretable and sound coherent with our intuition. Furthermore, the clusters revealed regular patterns of human daily activities that are consistent with previous findings regarding the strong predictability and regularity of human mobility.
We hope that our methodology will be helpful in future applications such as urban and transportation planning, the sociology of mobility behavior, and spreading dynamics. 

In future work, we plan to study the time dimension to propose novel indicators and analysis methods for semantic sequences in a \textit{time-structured approach}. 
For example, we could describe each activity according to its start and end timestamps.
Additionally, we hope to expand our methodology to account multidimensional semantic sequences. Integrating the time dimension and multidimensional semantics will facilitate the treatment of more detailed sequences and enhance our methodology.

\bibliographystyle{abbrv}        % 

\bibliography{ref_dataSem.bib}

\end{document}